\documentclass[10pt,reqno,draft]{amsart}
\usepackage{mathptmx}
\usepackage{helvet}
\usepackage{courier}
\usepackage[mathscr]{eucal}
\usepackage{amsthm}
\usepackage{amssymb}
\usepackage{enumerate}
\usepackage{pstricks}
\usepackage[amssymb]{SIunits}
\input{misdat.sty}
\input{misdat.env}
\begin{document}
\title{Updating beliefs with incomplete observations}
\author{Gert de Cooman}
\address{Ghent University, SYSTeMS Research Group,
Technologiepark -- Zwijnaarde 914, 9052 Zwijnaarde, Belgium}
\email{gert.decooman@ugent.be}
\author{Marco Zaffalon}
\address{IDSIA, Galleria 2, 6928 Manno (Lugano), Switzerland}
\email{zaffalon@idsia.ch}
\begin{abstract}
  Currently, there is renewed interest in the problem, raised by Shafer in
  1985, of updating probabilities when observations are incomplete (or
  set-valued). This is a fundamental problem in general, and of particular
  interest for Bayesian networks. Recently, Gr\"unwald and Halpern have shown
  that commonly used updating strategies fail in this case, except under very
  special assumptions.  In this paper we propose a new method for updating
  probabilities with incomplete observations. Our approach is deliberately
  conservative: we make no assumptions about the so-called incompleteness
  mechanism that associates complete with incomplete observations. We model
  our ignorance about this mechanism by a vacuous lower prevision, a tool from
  the theory of imprecise probabilities, and we use only coherence arguments
  to turn prior into posterior (updated) probabilities. In general, this new
  approach to updating produces lower and upper posterior probabilities and
  previsions (expectations), as well as partially determinate decisions.  This
  is a logical consequence of the existing ignorance about the incompleteness
  mechanism. As an example, we use the new updating method to properly address
  the apparent paradox in the `Monty Hall' puzzle. More importantly, we apply
  it to the problem of classification of new evidence in probabilistic
  expert systems, where it leads to a new, so-called \emph{conservative
    updating rule}. In the special case of Bayesian networks constructed using
  expert knowledge, we provide an exact algorithm for classification based on
  our updating rule, which has linear-time complexity for a class of networks
  wider than polytrees.  This result is then extended to the more general
  framework of credal networks, where computations are often much harder than
  with Bayesian nets. Using an example, we show that our rule appears to
  provide a solid basis for reliable updating with incomplete observations,
  when no strong assumptions about the incompleteness mechanism are justified.
\end{abstract}
\keywords{Incomplete observations, updating probabilities, imprecise
  probabilities, coherent lower previsions, vacuous lower previsions, naive
  updating, conservative updating, Bayesian networks, credal networks,
  puzzles, credal classification.}  \maketitle

\section{Introduction}
Suppose you are given two Boolean random variables, $C$ and $A$. $C=1$
represents the presence of a disease and $A=1$ is the positive result of a
medical test. You know that $p(C=0,A=0)=0.99$ and that $p(C=1,A=1)=0.01$, so
the test allows you to make a sure diagnosis.  However, it may happen that,
for some reason, the result of the test is missing. What should your diagnosis
be in this case? You might be tempted to say that the posterior probability of
$C=0$, conditional on a missing value of $A$, is simply $p(C=0\vert
A\in\{0,1\})=p(C=0)=0.99$, and that the diagnosis is `no disease' with high
probability. After all, this looks like a straightforward application of
Kolmogorov's \emph{definition} of conditional probability, which appears in
many textbooks: $\pr(B\vert E)=\pr(B\cap E)/\pr(E)$, for generic events $B$
and $E$, with $\pr(E)>0$.
\par
Unfortunately, it turns out that the above inference is wrong unless a
condition known in the literature as \emph{MAR} (\emph{missing at random}) is
satisfied.  MAR states that the probability that a measurement for $A$ is
missing, is the same both when conditioned on $A=0$ and when conditioned on
$A=1$, or, in other words, that there is no systematic reason for the missing
values of $A$ \cite{little1987}.
\par
The example above is a special case of the more general problem of updating
probabilities with observations that are \emph{incomplete}, or set-valued: it
could be argued that the fact that a measurement for $A$ is missing
corresponds to a set-valued observation of $\{0,1\}$ for $A$ rather than the
\emph{complete} or point-valued observations $0$ or $1$. The difficulty we are
facing is then how to update $p$ with such incomplete observations.  To our
knowledge, this problem was given serious consideration for the first time in
1985 by Shafer \cite{shafer1985}. Rather than taking traditional conditioning
as a definition, Shafer derived it from more primitive notions showing that
the right way to update probabilities with incomplete observations requires
knowledge of what we shall call the \emph{incompleteness mechanism} (called
\emph{protocol} in Shafer's paper), i.e., the mechanism that is responsible
for turning a complete observation into an incomplete one.  Shafer's result
tells us that neglecting the incompleteness mechanism leads to a naive
application of conditioning (also called \emph{naive conditioning} or
\emph{naive updating} in the following) that is doomed to failure in general.
This is evident when one addresses well-known puzzles by naive conditioning,
such as the three prisoners problem and the Monty Hall puzzle. What the
implications are in practise for more realistic applications of probability
theory, was partially addressed by Shafer when he observed that ``we do not
always have protocols in practical problems.'' In the example above, for
instance, we may not know which is the probability that a measurement $A$ is
missing conditional on $A=0$ and conditional on $A=1$ (such a conditional
probability is a specification of the protocol, or incompleteness mechanism).
We may not even know whether the two probabilities are equal \dots
\par
Surprisingly, Shafer's thesis seems to have been largely overlooked for many
years.\footnote{But see the discussion in \cite[Section~6.11]{walley1991},
  which has been a source of inspiration for the present work; and some papers
  by Halpern \emph{et al.} \cite{halpern1998,halpern1993}.} Kolmogorov's
influential formalisation of probability theory \cite{kolmogorov1950} may have
contributed in this respect: the way the definition of conditioning is
presented seems to suggest that one may be totally uninformed about the
incompleteness mechanism, and still be allowed to correctly update beliefs
after receiving some evidence $E$. That is, it seems to suggest that naive
updating is always the correct way to update beliefs. Actually, the definition
produces correct results when MAR does not hold only if the underlying possibility space is built in
such a way as to also model the incompleteness mechanism. Apart from the
influence of Kolmogorov's formalisation, we might identify the unclear
practical implications of Shafer's work as another reason for its being
considered by many as something of a statistical curiosity.
\par
The situation has changed recently, when an interesting paper by Gr\"unwald
and Halpern \cite{gruenwald2003} kindled a renewed interest in the subject. In
that work, strong arguments are presented for the following two theses: (i)
the incompleteness mechanism may be unknown, or difficult to model; and (ii)
the condition of \emph{coarsening at random} (or CAR \cite{gill1997}, a
condition more general than MAR), which guarantees that naive updating
produces correct results, holds rather infrequently. These two points taken
together do raise a fundamental issue in probability theory, which also
presents a serious problem for applications: how should one update beliefs
when little or no information is available about the incompleteness mechanism?
\par
In the above example, the mechanism might very well be such that $A$ cannot be
observed if and only if $A$ has the value $0$, and then $C=0$ would be a
certain conclusion. But it might equally well be the case that $A$ cannot be
observed if $A=1$, in which case $C=1$ would be certain. Of course, all the
intermediate randomised cases might also be possible. It follows that the
posterior probability of $C=0$ can, for all we know, lie anywhere in the
interval $[0,1]$, and our ignorance does not allow us to say that one value is
more likely than another. In other words, this probability is \emph{vacuous}.
Thus, knowing that the value of $A$ is missing, produces complete ignorance
about this probability and, as a result, total indeterminacy about the
diagnosis: we have no reason to prefer $C=0$ over $C=1$, or \emph{vice versa}.
All of this is a necessary and logical consequence of our ignorance about the
incompleteness mechanism. We cannot get around this indeterminacy, unless we
go back to the medical test and gather more relevant information about how it
may produce missing values.
\par
Generally speaking, we believe that the first step to answer the question
above is to recognise that there may indeed be ignorance about the
incompleteness mechanism, and to allow for such ignorance in our models. This
is the approach that we take in this paper. In
Section~\ref{sec:incomplete-observations}, we make our model as conservative
as possible by representing the ignorance about the incompleteness mechanism
by a \emph{vacuous lower prevision}, a tool from the theory of imprecise
probabilities \cite{walley1991}. Because we are aware that readers may not be
familiar with imprecise probability models, we present a brief discussion in
Section~\ref{sec:imprecise-probability}, with pointers to the relevant
literature.\footnote{See also \cite{walley1996} for a gentle and less dense
  introduction to imprecise probabilities with emphasis on artificial
  intelligence.} Loosely speaking, the vacuous lower prevision is equivalent
to the set of all distributions, i.e., it makes all incompleteness mechanisms
possible \emph{a priori}. Our basic model follows from this as a necessary
consequence, using the rationality requirement of \emph{coherence}. This
coherence is a generalisation to its imprecise counterpart of the requirements
of rationality in precise, Bayesian, probability theory \cite{finetti19745}.
We illustrate how our basic model works by addressing the Monty Hall puzzle,
showing that the apparent paradox vanishes if the knowledge that is actually
available about the incompleteness mechanism is modelled properly.
\par
We then apply our method for dealing with incomplete observations to the
special case of a classification problem, where objects are assigned to
classes on the basis of the values of their attributes.  The question we deal
with in Section~\ref{sec:missing-data}, is how classification should be done
when values for some of the attributes are missing. We derive a new updating
rule that allows us to deal with such missing data without making unwarranted
assumptions about the mechanism that produces these missing values. We regard
this so-called \emph{conservative updating rule} as a significant step toward
a general solution of the updating problem. Our rule leads to an imprecise
posterior, and as we argued above, it may lead to inferences that are
partially indeterminate. It may for instance happen that, due to the fact that
certain of the attribute values are missing, our method will assign an object
to a number of (optimal) classes, rather than to a single class, and that it
does not express any preference between these optimal classes. This
generalised way of doing classification is also called \emph{credal
  classification} in \cite{zaffalon2000}. As we have argued above, we have to
accept that this is the best our system can do, given the information that is
incorporated into it. If we want a more precise classification, we shall have
to go back and find out more about the mechanism that is responsible for the
fact that some attributes are missing. But, given the characteristics of our
approach, any such additional information will lead to a new classification
that refines ours, but can never contradict it, i.e., assign an object to a
class that was not among our optimal classes in the first place.
\par
In Section~\ref{sec:bayesian-nets}, we then apply the updating rule for
classification problems to Bayesian networks. We regard a Bayesian net as a
tool that formalises expert knowledge and is used to classify new evidence,
i.e., to select certain values of a class variable given evidence about the
attribute values. We develop an exact algorithm for credal classification with
Bayesian nets that is linear in the size of the input, when
the class node together with its Markov blanket is a singly connected graph.
Extension to the general case is provided by an approach analogous to
\emph{loop cutset conditioning}.  Section~\ref{sec:example} applies the
algorithm to an artificial problem and clarifies the differences with naive
updating. There are two important implications of the algorithmic complexity
achieved with Bayesian nets: the algorithm makes the new rule immediately
available for applications; and it shows that it is possible for the power of
robust, conservative, modelling to go hand in hand with efficient computation,
even for some multiply connected networks. This is enforced by our next
result: the extension of the classification algorithm to \emph{credal
  networks}, in Section~\ref{sec:credal nets}, with the same complexity.
Credal networks are a convenient way to specify partial prior knowledge. They
extend the formalism of Bayesian networks by allowing a specification in terms
of sets of probability measures. Credal nets allow the inherent imprecision in
human knowledge to be modelled carefully and expert systems to be developed
rapidly.  Such remarkable advantages have been partially overshadowed so far
by the computational complexity of working in the more general framework of
credal nets.  Our result shows that in many realistic scenarios, the
computational effort with credal networks is the same as that required by
Bayesian nets. This may open up a wealth of potential applications for credal
networks.
\par
The concluding Section~\ref{sec:conclusions} discusses directions and open
issues for future research. Additional, technical results have been gathered
in the appendices.

\section{Basic notions from the theory of imprecise probabilities}
\label{sec:imprecise-probability}
The theory of coherent lower previsions (sometimes also called the theory of
\emph{imprecise probabilities}\footnote{Other related names found in the
  literature are: indeterminate probabilities, interval (or interval-valued)
  probabilities, credal sets, \dots}) \cite{walley1991} is an extension of the
Bayesian theory of (precise) probability \cite{finetti1937,finetti19745}. It
intends to model a subject's uncertainty by looking at his dispositions toward
taking certain actions, and imposing requirements of rationality, or
consistency, on these dispositions.
\par
To make this more clear, consider a random variable $\rvx$ that may take
values in a finite\footnote{For simplicity, we shall only deal with variables
  with a \emph{finite} number of possible values in this paper.} set
$\xvalues$. A \emph{gamble} $f$ on the value of $\rvx$, or more simply, a
gamble on $\xvalues$, is a real-valued function on $\xvalues$. It associates a
(possibly negative) reward\footnote{In order to make things as simple as
  possible, we shall assume that these rewards are expressed in units of some
  predetermined \emph{linear} utility.} $f(x)$ with any value $x$ the random
variable $\rvx$ may assume. If a subject is uncertain about what value $\rvx$
assumes in $\xvalues$, he will be disposed to accept certain gambles, and to
reject others, and we may model his uncertainty by looking at which gambles he
accepts (or rejects).
\par
In the Bayesian theory of uncertainty (see for instance \cite{finetti19745}),
it is assumed that a subject can always specify a \emph{fair price}, or
\emph{prevision}, $\pr(f)$ for $f$, whatever the information available to him.
$\pr(f)$ is the unique real number such that the subject (i) accepts the
gamble $f-p$, i.e., accepts to buy the gamble $f$ for a price $p$, for all
$p<\pr(f)$; and (ii) accepts the gamble $q-f$, i.e., accepts to sell the
gamble $f$ for a price $q$, for all $q>\pr(f)$. In other words, it is assumed
that for essentially any real number $r$, the available information allows the
subject to decide which of the following two options he prefers: buying $f$
for price $r$, or selling $f$ for that price.
\par
It has been argued extensively \cite{smith1961,walley1991} that, especially if
little information is available about $\rvx$, there may be prices $r$ for
which a subject may have no real preference between these two options, or in
other words, that on the basis of the available information he remains
\emph{undecided} about whether to buy $f$ for price $r$ or to sell it for that
price: he may not be disposed to do either. If, as the Bayesian theory
requires, the subject \emph{should} choose between these two actions, his
choice will then not be based on any real preference: it will be arbitrary,
and not a realistic reflection of the subject's dispositions, based on the
available information.

\subsection{Coherent lower and upper previsions}
\begin{figure}[htbp]
  \centering \input{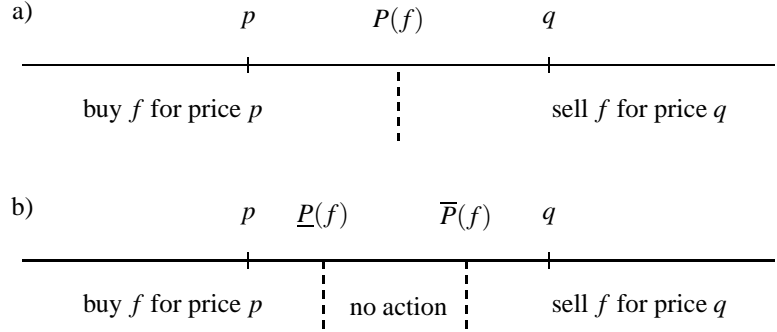}
\caption{Buying and selling a gamble $f$ in (a) the Bayesian theory,
  and (b) in imprecise probability theory}
\label{fig:bayesian-versus-imprecise}
\end{figure}

The theory of imprecise probabilities remedies this by allowing a subject to
specify two numbers: $\lpr(f)$ and $\upr(f)$.  The subject's \emph{lower
  prevision} $\lpr(f)$ for $f$ is the greatest real number $p$ such that he is
disposed to buy the gamble $f$ for all prices strictly smaller than $p$, and
his \emph{upper prevision} $\upr(f)$ for $f$ is the smallest real number $q$
such that he is disposed to sell $f$ for all prices strictly greater than $q$.
For any $r$ between $\lpr(f)$ and $\upr(f)$, the subject does not express a
preference between buying or selling $f$ for price $r$ (see
Figure~\ref{fig:bayesian-versus-imprecise}).
\par
Since selling a gamble $f$ for price $r$ is the same thing as buying $-f$ for
price $-r$, we have the following conjugacy relationship between lower and
upper previsions
\begin{equation}
\label{eq:conjugacy}
\upr(f)=-\lpr(-f).
\end{equation}
This tells us that whatever we say about upper previsions can always be
reformulated in terms of lower previsions. We shall therefore concentrate on
lower previsions. It will for the purposes of this paper suffice to consider
lower previsions $\lpr$ that are defined on the set $\gambles(\xvalues)$ of
all gambles on $\xvalues$, i.e., $\lpr$ is considered as a function that maps
any gamble $f$ on $\xvalues$ to the real number $\lpr(f)$.
\par
An \emph{event} $A$ is a subset of $\xvalues$, and it will be identified with
its \emph{indicator} $I_A$, which is a gamble assuming the value one on $A$
and zero elsewhere. We also denote $\lpr(I_A)$ by $\lpr(A)$ and call it the
\emph{lower probability} of the event $A$.  It is the supremum rate for which
the subject is disposed to bet on the event $A$. Similarly, the \emph{upper
  probability} $\upr(A)=\upr(I_A)=1-\lpr(\co A)$ is one minus the supremum
rate for which the subject is disposed to bet against $A$, i.e., to bet on the
complementary event $\co A$. Thus, events are special gambles, and lower and
upper probabilities are special cases of lower and upper previsions. We use
the more general language of gambles, rather than the more common language of
events, because Walley \cite{walley1991} has shown that in the context of
imprecise probabilities, the former is much more expressive and
powerful.\footnote{We shall see in Section~\ref{sec:linear-previsions} that
  for precise probabilities, both languages turn out to be equally
  expressive.}  For this reason, we consider `lower prevision' to be the
primary notion, and `lower probability' to be derived from it; and we follow
de Finetti's \cite{finetti19745} and Walley's \cite{walley1991} example in
using the same symbol $\pr$ for both (lower) \emph{p}revisions and (lower)
\emph{p}robabilities. Standard probabilistic practice would have us use the
symbols $\nex$ for \emph{e}xpectation/prevision and $\pr$ for
\emph{p}robability here.\footnote{Instead, we shall reserve the symbol $\nex$
  for natural extension.}
\par
Since lower previsions represent a subject's dispositions to act in certain
ways, they should satisfy certain criteria that ensure that these dispositions
are rational. \emph{Coherence} is the strongest such rationality requirement
that is considered in the theory of imprecise probabilities. For a detailed
definition and motivation, we refer to \cite{walley1991}. For the purposes of
the present discussion, it suffices to mention that a lower prevision $\lpr$
on $\gambles(\xvalues)$ is coherent if and only if it satisfies the following
properties, for all gambles $f$ and $g$ on $\xvalues$, and all non-negative
real numbers $\lambda$:
\begin{enumerate}[$(\lpr1)$]
\item $\min_{x\in\xvalues}f(x)\leq\lpr(f)$ [accepting sure gains];
\item $\lpr(f+g)\geq\lpr(f)+\lpr(g)$ [super-additivity];
\item $\lpr(\lambda f)=\lambda\lpr(f)$ [positive homogeneity].
\end{enumerate}
Observe that for a coherent $\lpr$, we have that $\upr(f)\geq\lpr(f)$ for all
$f\in\gambles(\xvalues)$.

\subsection{Linear previsions}
\label{sec:linear-previsions}
It follows from the behavioural interpretation of lower and upper previsions
that if $\lpr(f)=\upr(f)$ for some gamble $f$, then this common value is
nothing but the fair price, or prevision, $\pr(f)$ of $f$, as discussed in the
previous section. A \emph{linear prevision} $\pr$ on $\gambles(\xvalues)$ is
defined as a real-valued map on $\gambles(\xvalues)$ that is coherent when
interpreted as a lower prevision, and \emph{self-conjugate} in the sense that
$\pr(f)=-\pr(-f)$ for all gambles $f$, so the conjugate upper prevision of
$\pr$ is also given by $\pr$. This implies that a linear prevision $\pr$
should satisfy the following properties, for all gambles $f$ and $g$ on
$\xvalues$, and all real numbers $\lambda$:
\begin{enumerate}[$(\pr1)$]
\item $\min_{x\in\xvalues}f(x)\leq\pr(f)\leq\max_{x\in\xvalues}f(x)$;
\item $\pr(f+g)=\pr(f)+\pr(g)$;
\item $\pr(\lambda f)=\lambda\pr(f)$.
\end{enumerate}
This follows at once from the characterisation~$(\lpr1)$--$(\lpr3)$ of a
coherent lower prevision, and the conjugacy relationship~\eqref{eq:conjugacy}.
Thus, linear previsions turn out to be exactly the same thing as de Finetti's
coherent previsions \cite{finetti1937,finetti19745}. They are the so-called
\emph{precise} probability models, which turn out to be special cases of the
more general coherent imprecise probability models. Any linear prevision $\pr$
is completely determined by its so-called \emph{mass function} $p$, defined by
$p(x)=\pr(\{x\})$, since it follows from the axioms $(\pr2)$ and $(\pr3)$ that
for any gamble $f$,
\begin{equation*}
\pr(f)=\sum_{x\in\xvalues}f(x)p(x)
\end{equation*}
is the expectation of $f$ associated with the mass function $p$. We denote the
set of all linear previsions on $\gambles(\xvalues)$ by $\linprevs(\xvalues)$.

\subsection{Sets of linear previsions}\label{sec:credal sets}
With any lower prevision $\lpr$ on $\gambles(\xvalues)$, we can associate its
set of dominating linear previsions:
\begin{equation*}
\solp(\lpr)=\set{\pr\in\linprevs(\xvalues)}
{(\forall f\in\gambles(\xvalues))(\lpr(f)\leq\pr(f))}.
\end{equation*}
Observe that this set $\solp(\lpr)$ is convex and closed.\footnote{We only
  consider the topology of point-wise convergence on $\linprevs(\xvalues)$. If
  we identify linear previsions with their mass functions, which can in turn
  be identified with elements of the unit simplex in $\reals^n$, where $n$ is
  the cardinality of $\xvalues$, this topology is also the relativisation to
  this unit simplex of the usual Euclidean (metric) topology on $\reals^n$.}
It turns out that the lower prevision $\lpr$ is coherent if and only if
$\solp(\lpr)\not=\emptyset$, and if moreover $\lpr$ is the lower envelope of
$\solp(\lpr)$: for all gambles $f$ on $\xvalues$,\footnote{Since $\solp(\lpr)$
  is convex and closed, this infimum is actually achieved, and it can be
  replaced by a minimum.}
\begin{equation*}
\lpr(f)=\inf\set{\pr(f)}{\pr\in\solp(\lpr)}.
\end{equation*}
Conversely, the lower envelope $\lpr$ of any non-empty subset $\solp$ of
$\linprevs(\xvalues)$, defined by $\lpr(f)=\inf\set{\pr(f)}{\pr\in\solp}$ for
all $f\in\gambles(\xvalues)$, is a coherent lower prevision. Moreover
$\solp(\lpr)=\cc(\solp)$, where $\cc(\solp)$ is the convex closure (i.e., the
topological closure of the convex hull) of $\solp$ \cite[Chapters~2
and~3]{walley1991}. This tells us that working with coherent lower previsions
is equivalent to working with convex closed sets of linear previsions. It also
tells us that a coherent lower prevision $\lpr$ is also the lower envelope of
the set $\ext(\solp(\lpr))$ of the set of extreme points of $\solp(\lpr)$.
\par
This brings us to the so-called \emph{Bayesian sensitivity analysis
  interpretation} of a lower prevision $\lpr$ or a set of linear previsions
$\solp$. On this view, a subject's uncertainty should always be described by
some ideal probability measure, or equivalently, by some linear prevision
$\pr_T$. We could call this the \emph{assumption of ideal precision}. Due to
lack of time, resources or elicitation, we may not be able to uniquely
identify $\pr_T$, but we may often specify a set $\solp$ such that we are certain
that $\pr_T\in\solp$, or equivalently, a lower prevision $\lpr$ such that
$\lpr\leq\pr_T$. On this view, any conclusions or inferences we derive from
the available information must be \emph{robust}: they must be valid for all
possible candidates $\pr\in\solp$ for the ideal prevision $\pr_T$. Although we
emphatically do not make the assumption of ideal precision in this paper, we
shall see that many of the results we derive, are compatible with it, i.e.,
they can also be given a Bayesian sensitivity analysis interpretation.

\subsection{Vacuous lower previsions}
\label{sec:vacuous}
There is a class of coherent lower previsions that deserves special attention.
Consider a non-empty subset $B$ of $\xvalues$. Then the \emph{vacuous lower
  prevision $\lpr_B$ relative to $B$} is defined by
\begin{equation}
\label{eq:vacuous}
\lpr_B(f)=\min_{x\in B}f(x)
\end{equation}
for all gambles $f$ on $\xvalues$. Verify that $\lpr_B$ is a coherent lower
prevision, and moreover
\begin{equation*}
\solp(\lpr_B)=\set{\pr\in\linprevs(\xvalues)}{\pr(B)=1}.
\end{equation*}
This tells us that $\lpr_B$ is the smallest (and therefore most conservative)
coherent lower prevision $\lpr$ on $\gambles(\xvalues)$ that satisfies
$\lpr(B)=1$ (and therefore $\upr(B)=\pr(B)=1$).  $\lpr(B)=1$ means that it is
\emph{practically certain} to the subject that $\rvx$ assumes a value in $B$,
since he is disposed to bet at all non-trivial odds on this event. Thus, in
the context of the theory of lower probabilities, $\lpr_B$ is the appropriate
model for the piece of information that `$\rvx$ assumes a value in $B$'
\emph{and nothing more}: any other coherent lower prevision $\lpr$ that
satisfies $\lpr(B)=1$ dominates $\lpr_B$, and therefore represents stronger
behavioural dispositions than those required by coherence and this piece of
information alone. Also observe that
\begin{equation*}
\ext(\solp(\lpr_B))=\set{\pr_x}{x\in B},
\end{equation*}
where $\pr_x$ is the (degenerate) linear prevision on $\gambles(\xvalues)$ all
of whose probability mass lies in $x$, defined by $\pr_x(f)=f(x)$ for all
gambles $f$ on $\xvalues$. $\lpr_B$ is therefore the lower envelope of this
set of (degenerate) linear previsions, as is also apparent from
Eq.~\eqref{eq:vacuous}.

\subsection{Marginal lower previsions}
Now consider another random variable $\rvy$ that may assume values in a finite
set $\yvalues$. A coherent lower prevision $\lpr$ on $\gambles(\xyvalues)$ is
a model for a subject's uncertainty about the values that the joint random
variable $(\rvx,\rvy)$ assumes in $\xyvalues$. We can associate with $\lpr$
the so-called \emph{marginal} lower prevision $\lpr_\rvy$ on
$\gambles(\yvalues)$, defined as follows:
\begin{equation*}
\lpr_\rvy(g)=\lpr(g')
\end{equation*}
for all $g\in\gambles(\yvalues)$, where the gamble $g'$ on $\xyvalues$ is
defined by $g'(x,y)=g(y)$ for all $(x,y)\in\xyvalues$. In what follows, we
shall identify $g$ and $g'$, and simply write $\lpr(g)$ rather than
$\lpr(g')$. The marginal $\lpr_\rvx$ on $\gambles(\xvalues)$ is defined
similarly.
\par
The marginal $\lpr_\rvy$ is the corresponding model for the subject's
uncertainty about the value that $\rvy$ assumes in $\yvalues$, irrespective of
what value $\rvx$ assumes in $\xvalues$.
\par
If $\pr$ is in particular a linear prevision, its marginal $\pr_\rvy$ is a
linear prevision too, and its mass function $p_\rvy$ is given by the
well-known formula
\begin{equation*}
p_\rvy(y)=\pr(\xvalues\times\{y\})=\sum_{x\in\xvalues}p(x,y).
\end{equation*}

\subsection{Conditional lower previsions and separate coherence}
\label{sec:separate-coherence}
Consider any gamble $h$ on $\xyvalues$ and any value $y\in\yvalues$. A
subject's \emph{conditional lower prevision} $\lpr(h\vert\rvy=y)$, also
denoted as $\lpr(h\vert y)$, is the highest real number $p$ for which the
subject would buy the gamble $h$ for any price strictly lower than $p$, if he
knew in addition that the variable $\rvy$ assumes the value $y$ (and nothing
more!).
\par
We shall denote by $\lpr(h\vert\rvy)$ the \emph{gamble} on $\yvalues$ that
assumes the value $\lpr(h\vert\rvy=y)=\lpr(h\vert y)$ in $y\in\yvalues$. We
can for the purposes of this paper assume that $\lpr(h\vert\rvy)$ is defined
for all gambles $h$ on $\xyvalues$, and we call $\lpr(\cdot\vert\rvy)$ a
conditional lower prevision on $\gambles(\xyvalues)$.  Observe that
$\lpr(\cdot\vert\rvy)$ maps any gamble $h$ on $\xyvalues$ to the gamble
$\lpr(h\vert\rvy)$ on $\yvalues$.
\par
Conditional lower previsions should of course also satisfy certain rationality
criteria. $\lpr(\cdot\vert\rvy)$ is called \emph{separately coherent} if for
all $y\in\yvalues$, $\lpr(\cdot\vert y)$ is a coherent lower prevision on
$\gambles(\xyvalues)$, and if moreover $\lpr(\xvalues\times\{y\}\vert y)=1$.
This last condition is natural since it simply expresses that if the subject
knew that $\rvy=y$, he would be disposed to bet at all non-trivial odds on the
event that $\rvy=y$.
\par
It is a consequence of separate coherence that for all $h$ in
$\gambles(\xyvalues)$ and all $y\in\yvalues$,
\begin{equation*}
\lpr(h\vert y)=\lpr(h(\cdot,y)\vert y).
\end{equation*}
This implies that a separately coherent $\lpr(\cdot\vert\rvy)$ is completely
determined by the values $\lpr(f\vert\rvy)$ that it assumes in the gambles $f$
on $\xvalues$ alone. We shall use this very useful property repeatedly
throughout the paper.

\subsection{Joint coherence and the Generalised Bayes Rule}
\label{sec:joint-coherence}
If besides the (separately coherent) conditional lower prevision
$\lpr(\cdot\vert\rvy)$ on $\gambles(\xyvalues)$, the subject has also
specified a coherent (unconditional) lower prevision $\lpr$ on 
$\gambles(\xyvalues)$, then $\lpr$ and $\lpr(\cdot\vert\rvy)$ should in
addition satisfy the consistency criterion of \emph{joint coherence}. This
criterion is discussed and motivated at great length in
\cite[Chapter~6]{walley1991}. For our present purposes, it suffices to mention
that $\lpr$ and $\lpr(\cdot\vert\rvy)$ are jointly coherent if and only if
\begin{equation*}
\lpr(I_{\xvalues\times\{y\}}[h-\lpr(h\vert y)])=0\text{ for all } y\in\yvalues\text{ and all }h\in\gambles(\xyvalues).\tag{GBR}\label{GBR}
\end{equation*}
If $\lpr$ is a linear prevision $\pr$, this can be rewritten as
$\pr(hI_{\xvalues\times\{y\}})=\lpr(h\vert y)\pr(\xvalues\times\{y\})$, and if
$p_\rvy(y)=\pr_\rvy(\{y\})=\pr(\xvalues\times\{y\})>0$ it follows that
$\lpr(\cdot\vert y)$ is the precise (linear) prevision given by Bayes' rule:
\begin{equation*}
\lpr(h\vert y)=\pr(h\vert y)
=\dfrac{\pr(hI_{\xvalues\times\{y\}})}{\pr(\xvalues\times\{y\})},
\end{equation*}
or equivalently, in terms of mass functions: if $p_\rvy(y)>0$ then $p(x\vert
y)=p(x,y)/p_\rvy(y)$. For this reason, the joint coherence condition given
above is also called the \emph{Generalised Bayes Rule} (GBR, for short). It
can be shown \cite[Theorem~6.4.1]{walley1991} that if
$\lpr(\xvalues\times\{y\})>0$, then $\lpr(h\vert y)$ is uniquely determined by
this condition, or in other words: it is the unique solution of the following
equation in $\mu$:
\begin{equation*}
\lpr(I_{\xvalues\times\{y\}}[h-\mu])=0.
\end{equation*}
Equivalently, we then have that
\begin{equation*}
\lpr(h\vert y)
=\inf\set{\dfrac{\pr(hI_{\xvalues\times\{y\}})}
{\pr(\xvalues\times\{y\})}}{\pr\in\solp(\lpr)}, 
\end{equation*}
i.e., the uniquely coherent conditional lower prevision is obtained by
applying Bayes' rule to every linear prevision in $\solp(\lpr)$, and then
taking the lower envelope. For this reason, this procedure for obtaining a
conditional from a joint lower prevision is also called \emph{divisive
  conditioning} by Seidenfeld \emph{et al.}~\cite{herron1997,seidenfeld1993}.

\subsection{Natural and regular extension}
\label{sec:regular-extension}
If $\lpr(\xvalues\times\{y\})>0$, then the conditional lower prevision
$\lpr(\cdot\vert y)$ is uniquely determined by the unconditional lower
prevision $\lpr$. But this is no longer necessarily the case if
$\lpr(\xvalues\times\{y\})=0$ (something similar holds in the Bayesian theory
for precise previsions $\pr$ if $p_\rvy(y)=\pr(\xvalues\times\{y\})=0$). The
smallest, or most conservative, conditional lower prevision
$\nexl(\cdot\vert\rvy)$ that is jointly coherent with the joint lower
prevision $\lpr$ is called the \emph{natural extension} of $\lpr$ to a
conditional lower prevision. For any gamble $h$ on $\xyvalues$ and $y$ in
$\yvalues$, it is uniquely determined by the GBR if
$\lpr(\xvalues\times\{y\})>0$, and by
\begin{equation*}
\nexl(h\vert y)=\min_{x\in\xvalues}h(x,y)
\end{equation*}
if $\lpr(\xvalues\times\{y\})=0$, i.e., $\nexl(\cdot\vert y)$ is then the
\emph{vacuous} lower prevision relative to the set $\xvalues\times\{y\}$.
\par
In certain cases, it may be felt that natural extension is too conservative
when $\lpr(\xvalues\times\{y\})=0$. The following procedure, called
\emph{regular extension}, allows us to associate with any coherent lower
prevision $\lpr$ on $\gambles(\xyvalues)$ another (separately coherent)
conditional lower prevision $\rexl(\cdot\vert\rvy)$ that is jointly coherent
with $\lpr$:
\begin{enumerate}[(RE1)]
\item if $\upr(\xvalues\times\{y\})>0$, then $\rexl(h\vert y)$ is the greatest
  solution of the following inequality in $\mu$:
\begin{equation*}
\lpr\left(I_{\xvalues\times\{y\}}[h-\mu]\right)\geq0;
\end{equation*}
\item if $\upr(\xvalues\times\{y\})=0$, then $\rexl(\cdot\vert y)$ is the
  vacuous lower prevision relative to $\xvalues\times\{y\}$:
\begin{equation*}
\rexl(h\vert y)=\min_{x\in\xvalues}h(x,y);
\end{equation*}
\end{enumerate}
where $h$ is any gamble on $\xyvalues$. Regular extension coincides with
natural extension unless $\lpr(\xvalues\times\{y\})=0$ and
$\upr(\xvalues\times\{y\})>0$, in which case natural extension is vacuous and
regular extension can be much less conservative. We shall see examples of this
in the following sections.  The regular extension $\rexl(\cdot\vert\rvy)$ is
the smallest, or most conservative, conditional lower prevision that is
coherent with the joint $\lpr$ and satisfies an additional regularity
condition. It is the appropriate conditioning rule to use if a subject accepts
precisely those gambles $h$ for which $\lpr(h)\geq0$ and $\upr(h)>0$ (see
\cite[Appendix~J]{walley1991} for more details). It is especially interesting
because it has a nice interpretation in terms of sets of linear previsions: if
$\upr(\xvalues\times\{y\})>0$ it can be shown quite easily that
\begin{equation*}
\rexl(h\vert y)
=\inf\set{\dfrac{\pr(hI_{\xvalues\times\{y\}})}
{\pr(\xvalues\times\{y\})}}
{\text{$\pr\in\solp(\lpr)$ and $\pr(\xvalues\times\{y\})>0$}}.
\end{equation*}
Thus, $\rexl(h\vert y)$ can be obtained by applying Bayes' rule (whenever
possible) to the precise previsions in $\solp(\lpr)$, and then taking the
infimum. Regular extension therefore seems the right way to update lower
previsions on the Bayesian sensitivity analysis interpretation as well. It has
been called \emph{Bayesian updating} of coherent lower previsions by for
instance Jaffray \cite{jaffray1992}. Regular extension is also used for
updating in one of the more successful imprecise probability models, namely
Walley's Imprecise Dirichlet Model \cite{walley1996b}, where using natural
extension would lead to completely vacuous inferences. Also see
\cite{campos1990,fagin1991,walley1981,walley1991,walley1996} for more
information about this type of updating.

\subsection{Marginal extension}
\label{sec:marginal-extension}
It may also happen that besides a (separately coherent) conditional lower
prevision $\lpr(\cdot\vert\rvy)$ on $\gambles(\xyvalues)$ (or equivalently,
through separate coherence, on $\gambles(\xvalues)$), we also have a coherent
marginal lower prevision $\lpr_\rvy$ on $\gambles(\yvalues)$ modelling the
available information about the value that $\rvy$ assumes in $\yvalues$.
\par
We can then ask ourselves whether there exists a coherent lower prevision
$\lpr$ on all of $\gambles(\xyvalues)$ that (i) has marginal $\lpr_\rvy$, and
(ii) is jointly coherent with $\lpr(\cdot\vert\rvy)$. It turns out that this
is always possible. In fact, we have the following general theorem (a special
case of \cite[Theorem~6.7.2]{walley1991}), which is easily proved using the
results in the discussion above.

\begin{theorem}[Marginal extension theorem]
\label{theo:marginal-extension}
Let $\lpr_\rvy$ be a coherent lower prevision on $\gambles(\yvalues)$, and let
$\lpr(\cdot\vert\rvy)$ be a separately coherent conditional lower prevision on
$\gambles(\xyvalues)$. Then the smallest (most conservative) coherent lower
prevision on $\gambles(\xyvalues)$ that has marginal $\lpr_\rvy$ and that is
jointly coherent with $\lpr(\cdot\vert\rvy)$ is given by
\begin{equation}
\label{eq:marginal-extension}
\lpr(h)=\lpr_\rvy(\lpr(h\vert\rvy))
\end{equation}
for all gambles $h$ on $\xyvalues$.
\end{theorem}
\noindent
For a linear marginal $\pr_\rvy$ and a conditional linear prevision
$\pr(\cdot\vert\rvy)$, we again recover well-known results: the marginal
extension is the linear prevision $\pr=\pr_\rvy(\pr(\cdot\vert\rvy))$. In
terms of mass functions, the marginal extension of the marginal $p_\rvy(y)$
and the conditional $p(y\vert x)$ is given by $p(x,y)=p(x\vert y)p_\rvy(y)$.
Walley has shown \cite[Section~6.7]{walley1991} that marginal extension also
has a natural Bayesian sensitivity analysis interpretation in terms of sets of
linear previsions: for any gamble $h$ on $\xyvalues$, we have that
\begin{multline}
\label{eq:marginal-extension-robust}
\lpr(h)=\lpr_\rvy(\lpr(h\vert\rvy))\\
=\inf\set{\pr_\rvy(\pr(h\vert\rvy))} {\pr_\rvy\in\solp(\lpr_\rvy)\text{ and
  }(\forall y\in\yvalues) (\pr(\cdot\vert y)\in\solp(\lpr(\cdot\vert y)))}.
\end{multline}
The marginal extension of $\lpr_\rvy$ and $\lpr(\cdot\vert\rvy)$ can in other
words be obtained by forming the marginal extension for their compatible,
dominating linear previsions, and then taking the infimum.  In this infimum,
the sets $\solp(\lpr_\rvy)$ and $\solp(\lpr(\cdot\vert y))$ can be replaced by
their sets of extreme points.

\subsection{Decision making}
\label{sec:decision-making}
Suppose we have two actions $a$ and $b$, whose outcome depends on the actual
value that the variable $\rvx$ assumes in $\xvalues$. Let us denote by $f_a$
the gamble on $\xvalues$ representing the uncertain utility resulting from
action $a$: a subject who takes action $a$ receives $f_a(x)$ units of utility
if the value of $\rvx$ turns out to be $x$. Similar remarks hold for the
gamble $f_b$.
\par
If the subject is uncertain about the value of $\rvx$, it is not immediately
clear which of the two actions he should prefer.\footnote{Unless $f_a$
  point-wise dominates $f_b$ or \emph{vice versa}, which we shall assume is
  not the case.} But let us assume that he has modelled his uncertainty by a
coherent lower prevision $\lpr$ on $\gambles(\xvalues)$. Then he
\emph{strictly prefers} action $a$ to action $b$, which we denote as $a\spref
b$, if he is willing to pay some strictly positive amount in order to exchange
the (uncertain) rewards of $b$ for those of $a$. Using the behavioural
definition of the lower prevision $\lpr$, this can be written as
\begin{equation}
\label{eq:spref}
a\spref b\Leftrightarrow\lpr(f_a-f_b)>0.
\end{equation}
If $\lpr$ is a linear prevision $\pr$, this is equivalent to
$\pr(f_a)>\pr(f_b)$: the subject strictly prefers the action with the highest
expected utility. It is easy to see that $\lpr(f_a-f_b)>0$ can also be written
as
\begin{equation*}
(\forall\pr\in\solp(\lpr))(\pr(f_a)>\pr(f_b)).
\end{equation*}
In other words, $a\spref b$ if and only if action $a$ yields a higher expected
utility than $b$ for every linear prevision compatible with the subject's
model $\lpr$. This means that $\spref$ also has a reasonable Bayesian
sensitivity analysis interpretation. We shall say that a subject
\emph{marginally prefers} $a$ over $b$ if $\lpr(f_a-f_b)\geq0$, i.e., when he
is willing to exchange $f_b$ for $f_a$ in return for any strictly positive
amount of utility.
\par
If we now have some finite set of actions $K$, and an associated set of
uncertain rewards $\set{f_a}{a\in K}$, then it follows from the coherence of
the lower prevision $\lpr$ that the binary relation $\spref$ on $K$ is a
strict partial order, i.e., it is transitive and irreflexive. Optimal actions
$a$ are those elements of $K$ that are \emph{undominated}, i.e., to which no
other actions $b$ in $K$ are strictly preferred: $(\forall b\in K)(b\not\spref
a)$, or equivalently, after some manipulations,
\begin{equation*}
(\forall b\in K)(\upr(f_a-f_b)\geq0).
\end{equation*}
We shall call such actions \emph{$\lpr$-maximal} (in $K$). If $\lpr$ is a
linear prevision $\pr$, the $\pr$-maximal actions are simply those actions $a$
in $K$ with the highest expected utility $\pr(f_a)$.
\par
Two actions $a$ and $b$ are called \emph{equivalent} to a subject, which we
denote as $a\indif b$, if he is disposed to (marginally) exchange any of them
for the other, i.e., if both $\lpr(f_a-f_b)\geq0$ and $\lpr(f_b-f_a)\geq0$, or
equivalently,
\begin{equation*}
a\indif b
\Leftrightarrow
\upr(f_a-f_b)=\lpr(f_a-f_b)=\lpr(f_b-f_a)=\upr(f_b-f_a)=0.
\end{equation*}
When $\lpr$ is a linear prevision $\pr$, this happens precisely when
$\pr(f_a)=\pr(f_b)$, i.e., when both actions have the same expected utility.
\par
When $\lpr$ is imprecise, two actions $a$ and $b$ may be \emph{incomparable}:
they are neither equivalent, nor is either action strictly preferred over the
other. This happens when both $\lpr(f_a-f_b)\leq0$ and $\lpr(f_b-f_a)\leq0$
and at least one of these inequalities is strict. This means that the subject
has no preference (not even a marginal one) for one action over the other; he
is undecided. Note that this cannot happen for precise previsions.
\par
Any two $\lpr$-maximal actions are either equivalent (they always are when
$\lpr$ is precise), or incomparable, meaning that the information present in
the model $\lpr$ does not allow the subject to choose between them. It is an
essential feature of imprecise probability models that they allow for this
kind of indecision.

\section{Incomplete observations}
\label{sec:incomplete-observations}
We are now ready to describe our basic model for dealing with incomplete
observations. It is a general model that describes a situation where we want
to measure, or determine, the value of a certain variable $\srv$, but for some
reason can do so only in an imperfect manner: we perform some kind of
measurement whose outcome is $\orv$, but this does not allow us to completely
determine the value of $\srv$.
\par
Let us give a few concrete examples to make this more clear. Suppose we want
to measure the voltage ($\srv$) across a resistor, but the read-out ($\orv$)
of our digital voltage meter rounds this voltage to the next millivolt
(\milli\volt).  So if, say, we read that $\orv=12\milli\volt$, we only know
that the voltage $\srv$ belongs to the interval
$(11\milli\volt,12\milli\volt]$.
\par
In the example in the Introduction, $\srv=A$ is the result of the medical
test. If we know the result $x$ of the test, then we say that we observe
$\orv=x$. But if the test result is missing, we could indicate this by saying
that $\orv=\ast$ (or any other symbol to denote that we do not get a test
result $0$ or $1$). In that case, we only know that $\srv$ belongs to the set
$\{0,1\}$. 
\par
In the well-known \emph{three-prisoner problem}, three prisoners $a$, $b$ and
$c$ are waiting to be executed when it is decided that one of them, chosen
randomly, is to be set free. The warden tells prisoner $a$ the name of one of
the other two convicts, who has not been reprieved. The question is then if
what the warden tells $a$ gives him more information about whether he will be
executed or not. This can also be seen as a case of an incomplete observation:
the variable $\srv$ identifies which prisoner is to be reprieved, and the
observation $\orv$ is what the warden tells prisoner $a$. If for instance $a$
is reprieved, then the warden will name either $b$ or $c$: we then know that
$\orv$ can take any value in the set $\{b,c\}$. Conversely, if the warden
names prisoner $b$, so $\orv=b$, then all we know is that the variable $X$ can
take any value in $\{a,c\}$, so again $\srv$ is not completely determined by
the observation $\orv$. We shall see other concrete examples further in this
section as well as in the next section.
\par
Let us now present a formal mathematical model that represents the features
that are common to problems of this type. We consider a random variable $\srv$
that may assume values in a \emph{finite} set $\states$. Suppose that we have
some model for the available information about what value $\srv$ will assume
in $\states$, which takes the form of a coherent lower prevision $\lpr_0$
defined on $\gambles(\states)$.
\par
We now receive additional information about the value of $\srv$ by observing
the value that another random variable $\orv$ (the observation) assumes in a
\emph{finite} set of possible values $\observs$. Only, these observations are
\emph{incomplete} in the following sense: the value of $\orv$ does not allow
us to identify the value of $\srv$ uniquely. In fact, the only information we
have about the relationship between $\srv$ and $\orv$ is the following: if we
know that $\srv$ assumes the value $x$ in $\states$, then we know that $\orv$
must assume a value $o$ in a \emph{non-empty} subset $\mvm(x)$ of $\observs$,
\emph{and nothing more}. This idea of modelling incomplete observations
through a so-called \emph{multi-valued map} $\mvm$ essentially goes back to
Strassen \cite{strassen1964}.
\par
If we observe the value $o$ of $\orv$, then we know something more about
$\srv$: it can then only assume values in the set
\begin{equation*}
\{o\}^*=\set{x\in\states}{o\in\mvm(x)}
\end{equation*}
of those values of $\srv$ that \emph{may} produce the observation $\orv=o$. We
shall henceforth assume that $\{o\}^*\not=\emptyset$ for all $o\in\observs$:
observations $o$ for which $\{o\}^*=\emptyset$, cannot be produced by any $x$
in $\states$, and they can therefore be eliminated from the set $\observs$
without any further consequences.
\par
Unless $\{o\}^*$ is a singleton, the observation $\orv=o$ does not allow us to
identify a unique value for $\srv$; it only allows us to restrict the possible
values of $\srv$ to $\{o\}^*$. This is even the case if there is some possible
value of $X$ for which $o$ is the only compatible observation, i.e., if the
set
\begin{equation*}
\{o\}_*=\set{x\in\states}{\mvm(x)=\{o\}}
\end{equation*}
is non-empty: the set $\{o\}^*$ includes $\{o\}_*$ and may still contain more
than one element.
\par
The question we want to answer in this section, then, is how we can use this
new piece of information that $\orv=o$ to coherently update the prior lower
prevision $\lpr_0$ on $\gambles(\states)$ to a posterior lower prevision
$\lpr(\cdot\vert\orv=o)=\lpr(\cdot\vert o)$ on $\gambles(\states)$.
\par
In order to do this, we need to model the available information about the
relationship between $\srv$ and $\orv$, i.e., about the so-called
\emph{incompleteness mechanism} that turns the values of $\srv$ into their
incomplete observations $\orv$. In the special case that the marginal $\lpr_0$
is a (precise) linear prevision $\pr_0$ (with mass function $p_0$), it is
often assumed that this mechanism obeys the CAR condition, mentioned in the
Introduction:
\begin{equation}
\label{eq:car}
p(o\vert x)=p(o\vert y)>0
\tag{CAR}
\end{equation}
for all $o\in\observs$ and all $x$ and $y$ in $\{o\}^*$ such that $p_0(x)>0$
and $p_0(y)>0$ (see \cite{gill1997,gruenwald2003} for an extensive discussion
and detailed references). It is in other words assumed that the probability of
observing $\orv=o$ is not affected by the specific values $x$ of $X$ that may
actually lead to this observation $o$.  After a few manipulations involving
Bayes' rule, we derive from the CAR assumption that quite simply
\begin{equation}
\label{eq:car-result}
p(x\vert o)
=
\begin{cases}
  \dfrac{p_0(x)}{\pr_0(\{o\}^*)}=p_0(x\vert\{o\}^*)
  &\text{if $x\in\{o\}^*$}\\
  0 &\text{otherwise}.
\end{cases}
\end{equation}
This means that if we make the CAR assumption about the incompleteness
mechanism, then using the so-called \emph{naive updating
  rule}~\eqref{eq:car-result} is justified.
\par
For imprecise priors $\lpr_0$, this result can be generalised as follows for
observations $o$ such that $\lpr_0(\{o\}^*)>0$. Observe that
Theorem~\ref{theo:car} has an immediate Bayesian sensitivity analysis
interpretation.

\begin{theorem}
\label{theo:car}
Assume that $p(o\vert x)=p(o\vert y)>0$ for all $o\in\observs$ and all $x$ and
$y$ in $\{o\}^*$ such that $\upr_0(\{x\})>0$ and $\upr_0(\{y\})>0$. Let
$o\in\observs$ be such that $\lpr_0(\{o\}^*)>0$.  Then the conditional lower
prevision $\lpr(\cdot\vert o)$ is uniquely determined by coherence, and given
by
\begin{equation*}
\lpr(f\vert o)
=\inf\set{\frac{\pr(fI_{\{o\}^*})}{\pr(\{o\}^*)}}{\pr\in\solp(\lpr_0)}
=\inf\set{\pr(f\vert\{o\}^*)}{\pr\in\solp(\lpr_0)}
\end{equation*}
for all gambles $f$ on $\states$.
\end{theorem}

\begin{proof}
  Let $N=\set{x\in\states}{\upr_0(\{x\})=0}$. Then it follows from the
  coherence of $\lpr_0$ that $\upr_0(N)=0$. Moreover, for any gamble $f$ on
  $\states$, it follows from the coherence of $\lpr_0$ that
  $\lpr_0(f)=\lpr_0(fI_{\co N})$: $\lpr_0(f)$ only depends on the values that
  $f$ assumes outside $N$. Moreover, our generalised CAR assumption
  identifies, for all $x$ outside $N$, a conditional \emph{linear} prevision
  $\pr(\cdot\vert x)$ on $\gambles(\observs)$, and hence, by separate
  coherence, on $\gambles(\states\times\observs)$. We may therefore write,
  with some abuse of notation,\footnote{The abuse consists in assuming that
    the conditional lower previsions $\lpr(\cdot\vert x)$ are linear also for
    $x$ in $N$, which we can do because we have just shown that the value of
    the marginal extension does not depend on them.} for the marginal
  extension $\lpr$ of $\lpr_0$ and $\pr(\cdot\vert\srv)$:
\begin{equation*}
\lpr(h)=\lpr_0(\pr(h\vert\srv)),
\end{equation*}
for all gambles $h$ on $\states\times\observs$. It follows from coherence
arguments (see \cite[Section~6.7.3]{walley1991}) that $\lpr$ is the
\emph{only} joint lower prevision with marginal $\lpr_0$ that is jointly
coherent with $\pr(\cdot\vert\srv)$. It also follows readily from the
generalised CAR assumption that for the conditional mass function, $p(o\vert
x)=L_oI_{\{o\}^*}(x)$ for all $x$ outside $N$, where $L_o$ is some strictly
positive real number that only depends on $o$, not on $x$.  Consequently,
\begin{equation*}
\lpr(\states\times\{o\})
=\lpr_0(p(o\vert\srv))=\lpr_0(L_oI_{\{o\}^*})=L_o\lpr_0(\{o\}^*)>0,
\end{equation*}
where the inequality follows from the assumptions. It now follows from the
discussion in Sections~\ref{sec:joint-coherence}
and~\ref{sec:regular-extension} that $\lpr(\cdot\vert o)$ is uniquely
determined from the joint $\lpr$ by coherence, and given by
\begin{equation*}
\lpr(f\vert o)
=\inf\set{\dfrac{\pr(\pr(fI_{\states\times\{o\}}\vert\srv)}
{\pr(\pr(\states\times\{o\}\vert\srv))}}
{\text{$\pr\in\solp(\lpr_0)$ and
$\pr(\pr(\states\times\{o\}\vert\srv))>0$}}
\end{equation*}
for all gambles $f$ on $\states$. The proof is complete if we consider that
for all $\pr\in\solp(\lpr_0)$, $\pr(N)=0$, whence with obvious notations, also
using separate coherence,
\begin{multline*}
  \pr(\pr(fI_{\states\times\{o\}}\vert\srv)) =\sum_{x\in\states\setminus
    N}p(x)\pr(fI_{\states\times\{o\}}\vert x)
  =\sum_{x\in\states\setminus N}p(x)\pr(f(x)I_{\{o\}}\vert x)\\
  =\sum_{x\in\states\setminus N}p(x)f(x)p(o\vert x)
  =\sum_{x\in\states\setminus N}p(x)f(x)L_oI_{\{o\}^*}(x)
  =L_o\pr(fI_{\{o\}^*}),
\end{multline*}
and similarly
\begin{equation*}
\pr(\pr(\states\times\{o\}\vert\srv))
=\pr(p(o\vert\srv))
=L_o\pr(\{o\}^*)>0,
\end{equation*}
where the inequality follows from $\pr(\{o\}^*)\geq\lpr(\{o\}^*)>0$.
\end{proof}

However, Gr\"unwald and Halpern \cite{gruenwald2003} have argued convincingly
that CAR is a very strong assumption, which will only be justified in very
special cases.
\par
Here, we want to refrain from making such unwarranted assumptions in general:
we want to find out what can be said about the posterior
$\lpr(\cdot\vert\orv)$ if \emph{no} assumptions are made about the
incompleteness mechanism, apart from those present in the definition of the
multi-valued map $\mvm$ given above. This implies that anyone making
additional assumptions (such as CAR) about the incompleteness mechanism will
find results that are compatible but stronger, i.e., will find a posterior
(lower) prevision that will point-wise dominate ours.
\par
We proceed as follows. We have argued in Section~\ref{sec:vacuous} that the
appropriate model for the piece of information that `$O$ assumes a value in
$\mvm(x)$' is the vacuous lower prevision $\lpr_{\mvm(x)}$ on
$\gambles(\observs)$ relative to the set $\mvm(x)$. This means that we can
model the relationship between $X$ and $O$ through the following (vacuous)
conditional lower prevision $\lpr(\cdot\vert\srv)$ on $\gambles(\observs)$,
defined by
\begin{equation}
\label{eq:ideal-actual-first}
\lpr(g\vert x)=\lpr_{\mvm(x)}(g)=\min_{o\in\mvm(x)}g(o)
\end{equation}
for any gamble $g$ on $\observs$.  We have argued in
Section~\ref{sec:separate-coherence} that there is a unique separately
coherent conditional lower prevision that extends this to gambles on the space
$\states\times\observs$: for any gamble $h$ in
$\gambles(\states\times\observs)$,
\begin{equation}
\label{eq:ideal-actual}
\lpr(h\vert x)=\min_{o\in\mvm(x)}h(x,o).
\end{equation}
\par
Eq.~\eqref{eq:ideal-actual-first} also has an interesting Bayesian sensitivity
analysis interpretation. The coherent lower prevision $\lpr(\cdot\vert x)$ is
the lower envelope of the set
\begin{equation*}
\solp(\lpr(\cdot\vert x))
=\set{\pr(\cdot\vert x)}{\pr(\mvm(x)\vert x)=1}
\end{equation*}
of all linear previsions on $\gambles(\observs)$ that assign probability one
to the event $\mvm(x)$, i.e., for which it is certain that $\orv\in\mvm(x)$.
On the Bayesian sensitivity analysis interpretation, each such linear
prevision $\pr(\cdot\vert x)$ represents a so-called \emph{random
  incompleteness mechanism} (or a protocol, in Shafer's terminology
\cite{shafer1985}): a random mechanism that chooses an incomplete observation
$o$ from the set $\mvm(x)$ of observations compatible with state $x$, with
probability $p(o\vert x)$. The set $\solp(\lpr(\cdot\vert x))$ contains all
possible such random incompleteness mechanisms, and its lower envelope
$\lpr(\cdot\vert x)$ models that we have no information at all about which
random incompleteness mechanism is active.
\par
Using Walley's marginal extension theorem (see
Theorem~\ref{theo:marginal-extension} in
Section~\ref{sec:marginal-extension}), the smallest (unconditional) lower
prevision $\lpr$ on $\gambles(\states\times\observs)$ that extends $\lpr_0$
and is jointly coherent with the conditional lower prevision
$\lpr(\cdot\vert\srv)$ is given by
\begin{equation*}
\lpr(h)=\lpr_0(\lpr(h\vert\srv))
\end{equation*}
for all gambles $h$ on $\states\times\observs$.\footnote{See
  \cite{miranda2002} for a more general discussion with more mathematical
  detail.} In order to find the posterior lower prevision, we can now apply
the technique of regular extension, discussed in
Section~\ref{sec:regular-extension}. It yields the smallest (most
conservative) posterior lower prevision $\rexl(\cdot\vert\orv)$ that is
jointly coherent with $\lpr$ (and therefore with $\lpr_0$ and
$\lpr(\cdot\vert\states)$) and satisfies an additional regularity condition.
We have argued in Sections~\ref{sec:regular-extension}
and~\ref{sec:marginal-extension} that it also seems the right way to obtain a
posterior lower prevision on the Bayesian sensitivity analysis interpretation.

\begin{theorem}
\label{theo:regular-extension-simple}
Let $o\in\observs$ and let $f$ be any gamble on $\states$. If\/
$\upr_0(\{o\}^*)>0$, then
\begin{equation*}
\rexl(f\vert o)=\max\set{\mu}
{\lpr_0\left(I_{\{o\}_*}\max\{f-\mu,0\}
+I_{\{o\}^*}\min\{f-\mu,0\}\right)\geq0}.
\end{equation*}
If\/ $\upr(\{o\}^*)=0$ then $\rexl(f\vert o)=\min_{x\in\states}f(x)$.
\end{theorem}

\begin{proof}
  The discussion in Section~\ref{sec:regular-extension} tells us to look at
  the value of
  $\upr(\states\times\{o\})=\upr_0(\upr(\states\times\{o\})\vert\srv)$.
  Observe that for any $x\in\states$, by Eq.~\eqref{eq:ideal-actual},
\begin{equation*}
\upr(\states\times\{o\}\vert x)
=\max_{p\in\mvm(x)}I_{\states\times\{o\}}(x,p)=I_{\{o\}^*}(x),
\end{equation*}
whence $\upr(\states\times\{o\})\vert\srv)=I_{\{o\}^*}$ and consequently
$\upr(\states\times\{o\})=\upr_0(\{o\}^*)$.  If
$\upr(\states\times\{o\})=\upr_0(\{o\}^*)=0$ then the discussion in
Section~\ref{sec:regular-extension} tells us that $\rexl(\cdot\vert o)$ is
indeed the vacuous lower prevision on $\gambles(\states)$ (relative to the set
$\states$). If $\upr(\states\times\{o\})=\upr_0(\{o\}^*)>0$, then we know
that, by definition, $\rexl(f\vert o)$ is the greatest solution of the
following inequality in $\mu$:
\begin{equation*}
\lpr\left(I_{\states\times\{o\}}[f-\mu]\right)\geq0.
\end{equation*}
But for any $x\in\states$, we find that
\begin{align*}
  \lpr(I_{\states\times\{o\}}[f-\mu]\vert x)
  &=\min_{p\in\mvm(x)}I_{\states\times\{o\}}(x,p)[f(x)-\mu]\\
  &=\begin{cases}
    f(x)-\mu & \text{if $x\in\{o\}_*$}\\
    \min\{0,f(x)-\mu\} & \text{if $x\in\{o\}^*$ and $x\not\in\{o\}_*$}\\
    0 & \text{if $x\not\in\{o\}^*$}
\end{cases}\\
&=I_{\{o\}_*}(x)\max\{f(x)-\mu,0\}+I_{\{o\}^*}(x)\min\{f(x)-\mu,0\},
\end{align*}
whence indeed
\begin{equation*}
\lpr\left(I_{\states\times\{o\}}[f-\mu]\right)
=\lpr_0\left(I_{\{o\}_*}\max\{f-\mu,0\}
+I_{\{o\}^*}\min\{f-\mu,0\}\right).
\end{equation*}
This concludes the proof.
\end{proof}
\noindent
It also follows from this proof and the discussion in
Section~\ref{sec:regular-extension}, that the natural---as opposed to the
regular---extension $\nexl(\cdot\vert o)$ is vacuous whenever
$\lpr(\xvalues\times\{o\})=\lpr_0(\{o\}_*)=0$, and that $\nexl(h\vert o)$ is
the unique solution of the equation
\begin{equation*}
\lpr_0\left(I_{\{o\}_*}\max\{f-\mu,0\}+I_{\{o\}^*}\min\{f-\mu,0\}\right)=0 
\end{equation*}
in $\mu$ whenever $\lpr_0(\{o\}_*)>0$ (in which case regular and natural
extension coincide). We shall see later that there are interesting cases where
$\{o\}_*$ is empty, and where the natural extension $\nexl(\cdot\vert o)$ is
therefore the vacuous lower prevision relative to $\xvalues$. But this seems
needlessly imprecise, as we know from the observation $\orv=o$ that $\rvx$
should belong to the set $\{o\}^*$ of those values that can produce the
observation $o$, which may be a proper subset of $\xvalues$. We shall see in
Theorem~\ref{theo:marco-simple} that regular extension produces results that
are more intuitively acceptable in this respect.
\par
Let us now apply the results of Theorem~\ref{theo:regular-extension-simple} to
a puzzle of some standing in probability theory: the Monty Hall puzzle (see
for instance \cite{gruenwald2003} for further discussion and references). We
mention in passing that it is very closely related to the three prisoners problem, introduced at the beginning of the section, an that it can be dealt
with in an almost identical manner.

\subsection*{The Monty Hall puzzle}
In the Monty Hall game show, there are three doors. One of these doors leads
to a car, and the remaining doors each have a goat behind them.  You indicate
one door, and the show's host---let us call him Monty---now opens one of the
other doors, which has a goat behind it.  After this observation, should you
choose to open the door that is left, rather than the one you indicated
initially?
\par
To solve the puzzle, we reformulate it using our language of incomplete
observations. Label the doors from 1 to 3, and assume without loss of
generality that you picked door~1. Let the variable $\srv$ refer to the door
hiding the car, then clearly $\states=\{1,2,3\}$.  Observe that there is a
precise prior prevision $\pr_0$ determined by
$\pr_0(\{1\})=\pr_0(\{2\})=\pr_0(\{3\})=\frac{1}{3}$. The observation variable
$\orv$ refers to the door that Monty opens, and consequently
$\observs=\{2,3\}$ is the set of doors Monty can open. If the car is behind
door 1, Monty can choose between opening doors 2 and 3, so $\mvm(1)=\{2,3\}$,
and similarly, $\mvm(2)=\{3\}$ and $\mvm(3)=\{2\}$.  Since we know nothing at
all about how Monty will choose between the options open to him, we should
model the available information about the relation between $\srv$ and $\orv$
by the conditional lower prevision $\lpr(\cdot\vert\srv)$ given by
Eq.~\eqref{eq:ideal-actual}: for any gamble $h$ on $\statobs$,
\begin{equation*}
\lpr(h\vert 1)=\min\{h(1,2),h(1,3)\},\quad
\lpr(h\vert 2)=h(2,3),\quad
\lpr(h\vert 3)=h(3,2).
\end{equation*}
Applying the marginal extension theorem to the marginal $\pr_0$ and the
conditional lower prevision $\lpr(\cdot\vert\srv)$, we find the following
joint lower prevision $\lpr$ on $\gambles(\statobs)$:
\begin{equation*}
\lpr(h)
=\frac{1}{3}\min\{h(1,2),h(1,3)\}+\frac{1}{3}h(2,3)+\frac{1}{3}h(3,2),
\end{equation*}
for all gambles $h$ on $\statobs$.
\par
Assume without loss of generality that Monty opens door $2$. What can we say
about the updated lower prevision $\rexl(f\vert 2)$ when $f$ is any gamble on
$\states$? Since $\lpr(\states\times\{2\})=\frac{1}{3}>0$, we can use the GBR
to find the (uniquely!) coherent $\rexl(f\vert 2)$ as the unique solution of
the following equation in $\mu$:
\begin{equation*}
\lpr(I_{\states\times\{2\}}[f-\mu])
=\frac{1}{3}\min\{f(1)-\mu,0\}+\frac{1}{3}[f(3)-\mu]=0.
\end{equation*}
It is easy to see that
\begin{equation*}
\rexl(f\vert 2)
=\frac{1}{2}f(3)+\frac{1}{2}\min\{f(3),f(1)\}.
\end{equation*}
We are now ready to solve the puzzle. Which of the two actions should we
choose: stick to our initial choice and open door 1 (action $a$), or open door
3 instead (action $b$). In Table~\ref{table:monty-hall} we see the possible
outcomes of each action for the three possible values of $\srv$.
\begin{table}[htbp]
\centering
\begin{tabular}{c|ccc}
  & 1 & 2 & 3 \\\hline
$a$ & car & goat & goat\\
$b$ & goat & goat & car\\
$f_b-f_a$ & $-\Delta$ & $0$ & $\Delta$\\
\end{tabular}
\vspace{\baselineskip}
\caption{Possible outcomes in the Monty hall puzzle }
\label{table:monty-hall}
\end{table}
If the gamble $f_a$ on $\states$ represents the uncertain utility received
from action $a$, and similarly for $f_b$, then we are interested in the gamble
$f_b-f_a$, which represents the uncertain utility from exchanging action $a$
for action $b$. The possible values for this gamble are also given in
Table~\ref{table:monty-hall}, where $\Delta$ denotes the difference in utility
between a car and a goat, which is assumed to be strictly positive. Then we
find that
\begin{equation*}
\rexl(f_b-f_a\vert2)=
\frac{1}{2}\Delta+\frac{1}{2}\min\{\Delta,-\Delta\}=0
\end{equation*}
and
\begin{equation*}
\rexl(f_a-f_b\vert2)=
-\frac{1}{2}\Delta+\frac{1}{2}\min\{\Delta,-\Delta\}=-\Delta.
\end{equation*}
This implies that, with the notions and notations established in
Section~\ref{sec:decision-making}, $a\not\spref b$, $b\not\spref a$, and
$a\not\indif b$: the available information does not allow us to say which of
the two actions, sticking to door $1$ (action $a$) or choosing door $3$
(action $b$), is to be strictly preferred; and neither are these actions
equivalent. They are incomparable, and we should remain undecided on the basis
of the information available in the formulation of the puzzle.
\par
The same conclusion can also be reached in the following way. Suppose first
that Monty has decided on beforehand to always open door $3$ when the car is
behind door $1$. Since he has actually opened door $2$, the car cannot be
behind door $1$, and it must therefore be behind door $3$. In this case,
action $b$ is clearly strictly preferable to action $a$.  Next, suppose that
Monty has decided on beforehand to always open door $2$ when the car is behind
door $1$.  Since he actually opens door $2$, there are two equally likely
possibilities, namely that the car is behind door $1$ or behind door $3$. Both
actions $a$ and $b$ now have the same expected utility (zero), and none of
them is therefore strictly preferable to the other. Since both possibilities
are consistent with the available information, we cannot infer any (robust)
strict preference of one action over the other. A similar analysis was made by
Halpern \cite{halpern1998}.
\par
Observe that since $\rexl(f_b-f_a\vert2)=0$, you \emph{almost-prefer} $b$ to
$a$, in the sense that you are disposed to exchange $f_a$ for $f_b$ in return
for any strictly positive amount. In the slightly more involved case that
Monty could also decide not to open any door (denote this observation by $0$),
we now have $\observs=\{0,2,3\}$, $\mvm(1)=\{0,2,3\}$, $\mvm(2)=\{0,3\}$ and
$\mvm(3)=\{0,2\}$.  Consequently, $\{2\}_*=\emptyset$ and $\{2\}^*=\{1,3\}$,
and a similar analysis as before (see in particular
Theorem~\ref{theo:marco-simple} below) tells us that the updated lower
prevision is given by $\rexl(f\vert2)=\min\{f(1),f(3)\}$, and we get
$\rexl(f_b-f_a\vert2)=\rexl(f_a-f_b\vert2)=-\Delta$: now neither option is
even almost-preferred, let alone strictly preferred, over the other.

\subsection*{When naive updating is justified}
We are now in a position to take a closer look at the issue of when using the
naive updating rule~\eqref{eq:car-result} can be justified, even if nothing is
known about the incompleteness mechanism.
\par
We start with a precise prior prevision $\pr_0$ on $\gambles(\states)$ and
consider an incomplete observation $o\in\observs$. We shall assume that
$\{o\}_*$ is non-empty\footnote{If $\{o\}_*=\emptyset$ then the vacuous lower
  prevision $\lpr(\cdot\vert o)$ relative to $\states$ is coherent with the
  joint $\lpr$, and naive updating will not be justified, as it produces a
  precise posterior.} and that the mass function $p_0$ is strictly positive on
all elements of $\{o\}^*$. In this case, it follows from the discussion in
Section~\ref{sec:joint-coherence} and the proof of
Theorem~\ref{theo:regular-extension-simple} that the posterior lower prevision
after observing $o$ is \emph{uniquely} determined by coherence, and equal to
the regular extension $\rexl(\cdot\vert o)$.
\par
We shall see from the following discussion that using the naive posterior
$\pr_0(\cdot\vert\{o\}^*)$ is still justified, even if we know nothing at all
about the incompleteness mechanism, if and only if
\begin{equation}
\label{eq:car-justified}
\{o\}_*=\{o\}^*,
\tag{NAIVE-OK}
\end{equation}
i.e., if all the states that \emph{may} produce observation $o$ can
\emph{only} produce observation $o$.
\par
First of all, if~\eqref{eq:car-justified} holds, it follows immediately from
Theorem~\ref{theo:regular-extension-simple} and the assumptions that
\begin{equation*}
\rexl(f\vert o)
=\frac{\pr_0(fI_{\{o\}^*})}{\pr_0(\{o\}^*)}
=\pr_0(f\vert\{o\}^*),
\end{equation*}
indeed yielding the same result as naive updating does (see
Eq.~\eqref{eq:car-result}).
\par
We now show that~\eqref{eq:car-justified} is also necessary. If our regular
extension (and therefore coherence) produces the same result as naive updating
does, this implies that $\rexl(\cdot\vert o)$ is a linear prevision. So we
have that for any gamble $f$ on $\states$, $\rexu(f\vert o)=-\rexl(-f\vert
o)$. It then follows from Theorem~\ref{theo:regular-extension-simple}, after
some elementary manipulations, that for each gamble $f$ there is a unique
$\mu$ such that
\begin{multline*}
  \pr_0\left(I_{\{o\}_*}\max\{f-\mu,0\}
    +I_{\{o\}^*}\min\{f-\mu,0\}\right)\\
  =\pr_0\left(I_{\{o\}_*}\min\{f-\mu,0\} +I_{\{o\}^*}\max\{f-\mu,0\}\right)=0.
\end{multline*}
Let $x$ be any element of $\{o\}_*$. Choose in particular $f=I_{\{x\}}$, then
it follows that
\begin{equation*}
\pr_0\left(I_{\{o\}_*}[I_{\{x\}}-\mu]\right)
=\pr_0\left(I_{\{o\}^*}[I_{\{x\}}-\mu]\right)=0,
\end{equation*}
or equivalently
\begin{equation*}
\mu=\frac{p_0(x)}{\pr_0(\{o\}_*)}=\frac{p_0(x)}{\pr_0(\{o\}^*)},
\end{equation*}
whence $\pr_0(\{o\}_*)=\pr_0(\{o\}^*)$, since it follows from our assumptions
that $p_0(x)>0$. Again, since $p_0$ is assumed to be strictly positive on all
elements of $\{o\}^*$, Eq.~\eqref{eq:car-justified} follows.
\par
Observe that if Eq.~\eqref{eq:car-justified} holds, then all states $x$ in
$\{o\}^*$ can only lead to observation $o$, whence $p(o\vert x)=1$, so the CAR
condition is forced to hold, but in a very trivial way. In the same vein, it
follows from Eq.~\eqref{eq:car-justified} and Eq.~\eqref{eq:ideal-actual} that
for all $x$ in $\{o\}^*$, $\lpr(f\vert x)=f(o)$, so $\lpr(\cdot\vert x)$ is a
precise conditional prevision, whose mass function satisfies $p(o\vert x)=1$
for all $x$ in $\{o\}^*$.
\par
Our conclusion is that when the incompleteness mechanism is unknown,
\emph{naive updating is never justified}, except in those trivial situations
where CAR \emph{cannot} fail to hold. It is striking that Gr\"unwald and
Halpern obtain essentially the same conclusion using a rather different
approach: compare Eq.~\eqref{eq:car-justified} to Proposition~4.1 in
\cite{gruenwald2003}.

\subsection*{When an observation is not a necessary consequence}
To conclude this general discussion of incomplete observations, we shall
consider an important special case where nearly all reference to the prior is
obliterated\footnote{This is essentially due to the fact that updating
  requires us to condition on a set with zero lower prior probability. Observe
  that also in the case of precise probabilities, coherence imposes a very
  weak link between a prior and a posterior obtained after observing a set of
  zero prior probability. See also Section~\ref{sec:regular-extension}.} from
the posterior: we want to find $\rexl(\cdot\vert o)$ for an observation
$\orv=o$ that is not a necessary consequence of any value of $\srv$, i.e.,
\begin{equation}
\label{eq:ass1}
\{o\}_*=\set{x\in\states}{\mvm(x)=\{o\}}=\emptyset.
\tag{A1}
\end{equation}
We make the additional assumption that each state of the world compatible with
observation $o$ has positive upper probability, i.e.,
\begin{equation}
\label{eq:ass2}
\text{$\upr_0(\{x\})>0$ for all $x\in\{o\}^*$.}
\tag{A2}
\end{equation}
Under these conditions the regular extension $\rexl(\cdot\vert o)$ does not
depend on the prior $\lpr_0$, and only retains the information present in the
multi-valued map $\mvm$, as the following theorem states. We also want to
observe that using natural rather than regular extension here, would lead to a
posterior that is vacuous with respect to all of $\xvalues$, which would make
us lose even the information present in $\mvm$.

\begin{theorem}
\label{theo:marco-simple}
If\/ $o\in\observs$ satisfies Assumption~\eqref{eq:ass1} and $\lpr_0$
satisfies Assumption~\eqref{eq:ass2}, then $\rexl(\cdot\vert o)$ is the
vacuous lower prevision $\lpr_{\{o\}^*}$ on $\gambles(\states)$ relative to
$\{o\}^*$:
\begin{equation*}
\rexl(f\vert o)=\lpr_{\{o\}^*}(f)=\min_{x\colon o\in\mvm(x)}f(x)
\end{equation*}
for all $f$ in $\gambles(\states)$.
\end{theorem}

\begin{proof}
  We apply the results of Theorem~\ref{theo:regular-extension-simple}.  Since
  it follows from Assumption~\eqref{eq:ass2} and the coherence of $\lpr_0$
  that $\upr_0(\{o\}^*)>0$, we consider the gamble
\begin{equation*}
f_\mu
=I_{\{o\}^*}\min\{f-\mu,0\}+I_{\{o\}_*}\max\{f-\mu,0\}
=I_{\{o\}_*}\min\{f-\mu,0\}
\end{equation*}
on $\states$, where the last equality follows from Assumption~\eqref{eq:ass1}.
Then, we know that
\begin{equation*}
\rexl(f\vert o)
=\max\set{\mu}{\lpr_0(f_\mu)\geq0}
=\max\set{\mu}{\lpr_0(I_{\{o\}_*}\min\{f-\mu,0\})\geq0}.
\end{equation*}
Let $\lambda=\min_{x\colon o\in\mvm(x)}f(x)=\min_{x\in\{o\}^*}f(x)$.  If
$\mu\leq\lambda$ then $f(x)-\mu<0$ implies $f(x)-\lambda<0$ whence
$x\not\in\{o\}^*$.  Consequently $f_\mu$ is identically zero, whence
$\lpr_0(f_\mu)=0$. Assume therefore that $\mu>\lambda$. It remains to prove
that $\lpr_0(f_\mu)<0$. Observe that there is some $x_0$ in $\{o\}^*$ such
that $f(x_0)=\lambda$. If $f$ is constant, and therefore equal to $\lambda$,
on $\{o\}^*$, we find that $f_\mu=-[\mu-\lambda]I_{\{o\}^*}$, whence
\begin{equation*}
\lpr_0(f_\mu)=-[\mu-\lambda]\upr_0(\{o\}^*)<0,
\end{equation*}
also taking into account that Assumption~\eqref{eq:ass2} implies
$\upr_0(\{o\}^*)>0$. If $f$ is not constant on $\{o\}^*$, let $x_1$ be an
element of $\{o\}^*$ such that $f$ assumes no values between $f(x_0)$ and
$f(x_1)$ on $\{o\}^*$, and let $A_0=\set{x\in\{o\}^*}{f(x)=f(x_0)}$. Assume
that $\lambda<\mu<f(x_1)$, then for all $x\in\{o\}^*$ it follows from
$f(x)<\mu$ that $x\in A_0$ and therefore $f(x)=f(x_0)=\lambda$.  Consequently,
$f_\mu=-[\mu-\lambda]I_{A_0}$, whence
\begin{equation*}
\lpr_0(f_\mu)=-[\mu-\lambda]\upr_0(A_0)<0,
\end{equation*}
since it follows from Assumption~\eqref{eq:ass2} and the coherence of $\lpr_0$
that $\upr_0(A_0)>0$. Since we can also deduce from the coherence of $\lpr_0$
that $\lpr_0(f_\mu)$ is non-increasing in $\mu$, the result follows.
\end{proof}
\noindent
It is illustrative to prove this theorem in an alternative manner, using sets
of linear previsions.

\begin{proof}[Alternative proof using sets of linear previsions]
  A \emph{selection} $s$ for the multi-valued map $\mvm$ is a function from
  $\states$ to $\observs$ that associates with each $x\in\states$ a compatible
  observation $s(x)\in\mvm(x)$. Denote by $S(\mvm)$ the set of all possible
  selections:
\begin{equation*}
S(\mvm)=\set{s\in\observs^{\states}}
{(\forall x\in\states)(s(x)\in\mvm(x))}.
\end{equation*}
For any $s$ in $S(\mvm)$, define the conditional linear prevision
$\pr_s(\cdot\vert\srv)$ on $\gambles(\observs)$ by $\pr_s(\cdot\vert
x)=\pr_{s(x)}$ for all $x\in\states$, where $\pr_{s(x)}$ is the (degenerate)
linear prevision on $\gambles(\observs)$ all of whose probability mass lies in
$s(x)$, defined by $\pr_{s(x)}(g)=g(s(x))$ for all gambles $g$ on $\observs$.
Then clearly,
\begin{equation*}
\set{\pr_s(\cdot\vert\srv)}{s\in S(\mvm)}
\end{equation*}
is precisely the set of all conditional linear previsions
$\pr(\cdot\vert\srv)$ such that
\begin{equation*}
\pr(\cdot\vert x)\in\ext(\solp(\lpr(\cdot\vert x)))
\end{equation*}
for all $x\in\states$, and consequently, following the discussion in
Sections~\ref{sec:regular-extension} and~\ref{sec:marginal-extension}, it is
easily seen that
\begin{multline*}
  \rexl(f\vert o)\\
  =\inf\set{\dfrac{\pr_0(\pr_s(fI_{\states\times\{o\}}\vert\srv))}
    {\pr_0(\pr_s(\states\times\{o\}\vert\srv))}}
  {\text{$\pr_0\in\solp(\lpr_0)$, $s\in S(\mvm)$,
      $\pr_0(\pr_s(\states\times\{o\}\vert\srv))>0$}}.
\end{multline*}
Now for any $x$ in $\states$, also using separate coherence,
\begin{equation*}
\pr_s(\states\times\{o\}\vert x)
=I_{\states\times\{o\}}(x,s(x))=I_{\{o\}}(s(x))=I_{s^{-1}(\{o\})}(x),
\end{equation*}
whence $\pr_0(\pr_s(\states\times\{o\}\vert\srv))=\pr_0(s^{-1}(\{o\}))$, where
$s^{-1}(\{o\})=\set{x\in\states}{s(x)=o}\subseteq\{o\}^*$.  Similarly,
\begin{equation*}
\pr_s(fI_{\states\times\{o\}}\vert x)
=f(x)I_{\states\times\{o\}}(x,s(x))=f(x)I_{\{o\}}(s(x))
=f(x)I_{s^{-1}(\{o\})}(x),
\end{equation*}
whence $\pr_0(\pr_s(fI_{\states\times\{o\}}\vert\srv))
=\pr_0(fI_{s^{-1}(\{o\})})$. Consequently,
\begin{multline}
\label{eq:re-intermediate}
\rexl(f\vert o)
=\inf\set{\dfrac{\pr_0(fI_{s^{-1}(\{o\})})}{\pr_0(s^{-1}(\{o\}))}}
{\text{$\pr_0\in\solp(\lpr_0)$, $s\in S(\mvm)$,
    $\pr_0(s^{-1}(\{o\}))>0$}}\\
=\inf\set{\pr_0(f\vert s^{-1}(\{o\}))} {\text{$\pr_0\in\solp(\lpr_0)$, $s\in
    S(\mvm)$, $\pr_0(s^{-1}(\{o\}))>0$}}.
\end{multline}
Now consider any $x\in\{o\}^*$, whence $o\in\mvm(x)$. Consequently, there is a
selection $s\in S(\mvm)$ such that $s(x)=o$. Moreover,
Assumption~\eqref{eq:ass1} tells us that we can let $s(y)\not=o$ for all
$y\not=x$. Indeed, this is guaranteed if for all $y\not=x$ there is some $p$
in $\mvm(y)$ different from $o$, so that we can let $s(y)=p$.  If this
condition did not hold, then there would be some $y\not=x$ such that $p=o$ for
all $p\in\mvm(y)$, i.e., $\mvm(y)=\{o\}$, whence $y\in\{o\}_*$, which
contradicts Assumption~\eqref{eq:ass1}. Now for such $s$ it holds that
$s^{-1}(\{o\})=\{x\}$, and consequently $\pr_0(s^{-1}(\{o\}))=\pr_0(\{x\})$
and $\pr_0(fI_{s^{-1}(\{o\})})=f(x)\pr_0(\{x\})$ for all
$\pr_0\in\solp(\lpr_0)$. But Assumption~\eqref{eq:ass2} tells us that there is
at least one $\pr_0$ in $\solp(\lpr_0)$ for which $\pr_0(\{x\})>0$, and it
therefore follows from Eq.~\eqref{eq:re-intermediate} that $\rexl(f\vert
o)\leq f(x)$, and consequently $\rexl(f\vert o)\leq\min_{x\in\{o\}^*}f(x)$. To
prove the converse inequality, use Eq.~\eqref{eq:re-intermediate} and observe
that for all $s\in S(\mvm)$ and $\pr_0\in\solp(\lpr_0)$ such that
$\pr_0(s^{-1}(\{o\}))>0$,
\begin{equation*}
\dfrac{\pr_0(fI_{s^{-1}(\{o\})})}{\pr_0(s^{-1}(\{o\}))}
\geq\min_{x\in s^{-1}(\{o\})}f(x)
\geq\min_{x\in\{o\}^*}f(x),
\end{equation*}
since the left-hand side is some convex combination of the $f(x)$ for $x$ in
$s^{-1}(\{o\})$, and since $s^{-1}(\{o\})\subseteq\{o\}^*$.
\end{proof}

The selections $s\in S(\mvm)$ in this proof are essentially the deterministic
incompleteness mechanisms. They model that for any state $x$, the observation
$s(x)\in\mvm(x)$ is selected with probability one: $p_s(s(x)\vert x)=1$.

\section{Missing data in a classification problem}
\label{sec:missing-data}
In order to illustrate the practical implications of our model for the
incompleteness mechanism, let us show how it can be applied in classification
problems, where objects have to be assigned to a certain class on the basis of
the values of their attributes.

\subsection*{The basic classification problem}
Let in such a problem $\classes$ be the set of possible classes that we want
to assign objects to. Let $\attribs_1$, \dots, $\attribs_n$ be the sets of
possible values for the $n$ attributes on the basis of which we want to
classify the objects. We denote their Cartesian product by
\begin{equation*}
\states=\attribs_1\times\dots\times\attribs_n.
\end{equation*}
We consider a \emph{class variable} $C$, which is a random variable in
$\classes$, and \emph{attribute variables} $\arv_k$, which are random
variables in $\attribs_k$ ($k=1,\dots,n$).  The $n$-tuple
$\srv=(\arv_1,\dots,\arv_n)$ is a random variable in $\states$, and is called
the \emph{attributes variable}.  The available information about the
relationship between class and attribute variables is specified by a (prior)
lower prevision $\lpr_0$ on $\gambles(\classes\times\states)$, or
equivalently,\footnote{This is, provided that $\lpr_0(\classes\times\{x\})>0$
  for all $x\in\states$.} by a marginal lower prevision $\lpr_0$ on
$\gambles(\states)$ and a conditional lower prevision $\lpr_0(\cdot\vert\srv)$
on $\gambles(\classes)$.
\par
To see how classification is performed, let us first look at the case that
$\lpr_0$ is a linear prevision $\pr_0$, or equivalently, a precise probability
measure. If the attributes variable $\srv$ assumes a value $x$ in $\states$,
then the available information about the values of the class variable $C$ is
given by the conditional linear prevision $\pr_0(\cdot\vert x)$. If, on the
basis of the observed value $x$ of the attributes variable $\srv$, we decide
that some $c'$ in $\classes$ is the right class, then we can see this as an
action with an uncertain reward $f_{c'}$, whose value $f_{c'}(c)$ depends on
the value $c$ that $\crv$ actually assumes. An \emph{optimal class}
$c_{\mathrm{opt}}$ is one that maximises the expected reward
$\pr_0(f_{c'}\vert x)$: $\pr_0(f_{c_{\mathrm{opt}}}\vert
x)\geq\pr_0(f_{c'}\vert x)$ for all $c'\in\classes$.  As a common example, if
we let $f_{c'}=I_{\{c'\}}$, then $\pr_0(f_{c'}\vert x)=p_0(c'\vert x)$, and
this procedure associates the most probable class with each value $x$ of the
attributes.
\par
How can this be generalised to the more general case that $\lpr_0$ is not a
linear prevision?  If the attributes variable $\srv$ assumes a value $x$ in
$\states$, then the available information about the values of the class
variable $C$ is given by the conditional lower prevision $\lpr_0(\cdot\vert
x)$. The discussion in Section~\ref{sec:decision-making} then tells us that
the lower prevision $\lpr_0(\cdot\vert x)$ induces a strict preference
$\spref$ on the set of classes $\classes$ by
\begin{equation*}
c'\spref c''
\Leftrightarrow
\lpr_0(f_{c'}-f_{c''}\vert x)>0.
\end{equation*}
An optimal class $c_{\mathrm{opt}}$ is now one that is \emph{undominated},
i.e., such that for all $c'\in\classes$:
\begin{equation*}
\upr_0(f_{c_{\mathrm{opt}}}-f_{c'}\vert x)\geq0.
\end{equation*}
Observe that this reduces to the previously mentioned maximum expected utility
condition $\pr_0(f_{c_{\mathrm{opt}}}\vert x)\geq\pr_0(f_{c'}\vert x)$ when
$\lpr_0(\cdot\vert x)$ is a precise, or linear, prevision.
\par
To make this more clear, let us consider a medical domain, where
classification is used to make a diagnosis. In this case, the classes are
possible diseases and each attribute variable represents a measure with random
outcome. For example, attribute variables might represent medical tests, or
information about the patient, such as age, gender, life style, etc. We can
regard the specific instance of the vector of attribute variables for a
patient as a profile by which we characterise the person under examination.
The relationship between diseases and profiles is given by a joint mass
function on the class and the attribute variables. This induces a linear
prevision $\pr_0$ on $\gambles(\classes\times\states)$, according to
Section~\ref{sec:imprecise-probability}. A diagnosis is then obtained by
choosing the most probable disease given, or conditional on, a profile.
\par
In the case of a linear, or precise, $\pr_0(\cdot\vert x)$, if there is more
than one optimal class, all these classes are equivalent, as they have the
same expected reward.  But as we have explained in
Section~\ref{sec:decision-making}, this is no longer necessarily so for
imprecise $\lpr_0(\cdot\vert x)$.  Among the optimal, undominated classes,
there may be classes $c'$ and $c''$ that are not equivalent but
\emph{incomparable}: the information in $\lpr_0(\cdot\vert x)$ does not allow
us to choose between $c'$ and $c''$, and for all we know, both are possible
candidates for the class that the object is assigned to. This implies that if
we classify using an imprecise model $\lpr_0(\cdot\vert x)$, the best we can
often do, is assign a \emph{set} of possible, optimal classes to an object
with attributes $x$. In our medical example, a given profile would then lead
to a number of optimal candidate diagnoses, none of which is considered to be
better than (or even as good as) the others. Classifiers that allow for such
set-valued classification are called \emph{credal classifiers}
\cite{zaffalon2000}.

\subsection*{Dealing with missing data}
Now it may also happen that for a patient some of the attribute variables
cannot be measured, i.e., they are missing, e.g., when for some reason a
medical test cannot be done. In this case the profile is incomplete and we can
regard it as the set of all the complete profiles that are consistent with it.
As the above classification procedure needs profiles to be complete, the
problem that we are now facing, is how we should update our confidence about
the possible diseases given a set-profile.
\par
In more general terms, we observe or measure the value $a_k$ of some of the
attribute variables $\arv_k$, but not all of them. If a measurement is lacking
for some attribute variable $\arv_\ell$, it can in principle assume any value
in $\attribs_\ell$. This means that we can associate with any attribute
variable $\arv_k$ a so-called \emph{observation variable} $\orv_k$. This is a
random variable taking values in the set
\begin{equation*}
\observs_k=\attribs_k\cup\{\ast\},
\end{equation*}
whose elements are either the possible values of $\arv_k$, or a new element
$\ast$ which denotes that the measurement of $\arv_k$ is missing.
\par
Attribute variables $\arv_k$ and their observations $\orv_k$ are linked in the
following way: with each possible value $a_k\in\attribs_k$ of $\arv_k$ there
corresponds the following set of corresponding possible values for $\orv_k$:
\begin{equation}
\label{eq:md-mvm}
\mvm_k(a_k)=\{a_k,\ast\}\subseteq\observs_k.
\end{equation}
This models that whatever value $a_k$ the attribute variable $\arv_k$ assumes,
there is some mechanism, called the \emph{missing data mechanism}, that either
produces the (exact) observation $a_k$, or the observation $\ast$, which
indicates that a value for $\arv_k$ is missing. For the attributes variable
$\srv$ we then have that with each possible value $x=(a_1,\dots,a_n)$ there
corresponds a set of corresponding possible values for the \emph{observations
  variable} $\orv=(\orv_1,\dots,\orv_n)$:
\begin{equation*}
\mvm(x)=\mvm_1(a_1)\times\dots\times\mvm_n(a_n)\subseteq\observs,
\end{equation*}
where $\observs=\observs_1\times\dots\times\observs_n$. To summarise, we have
defined a multi-valued map $\mvm\colon\states\to\wp(\observs)$, whose
interpretation is the following: if the actual value of the attributes
variable $\srv$ is $x$, then due to the fact that, for some reason or another,
measurements for some attributes may be missing, the observations $\orv$ must
belong to $\mvm(x)$.
\par
So, in general, we observe some value $o=(o_1,\dots,o_n)$ of the variable
$\orv$, where $o_k$ is either the observed value for the $k$-th attribute, or
$\ast$ if a value for this attribute is missing.  In order to perform
classification, we therefore need to calculate a coherent updated lower
prevision $\lpr(\cdot\vert\orv=o)$ on $\gambles(\classes)$. This is what we
now set out to do.
\par
In order to find an appropriate updated lower prevision $\lpr(\cdot\vert o)$,
we need to model the available information about the relationship between
$\srv$ and $\orv$, i.e., about the missing data mechanism that produces
incomplete observations $\orv$ from attribute values $\srv$.
\par
We have arrived at a special case of the model described in the previous
section, and our so-called missing data mechanism is a particular instance of
the incompleteness mechanism described there. In this special case, it is easy
to verify that the general CAR assumption, discussed previously, reduces to
what is known in the literature as the MAR assumption \cite{little1987}: the
probability that values for certain attributes are missing, is not affected by
the specific values that these attribute variables assume.  MAR finds
appropriate justification in some statistical applications, e.g., special
types of survival analysis. However, there is strongly motivated criticism
about the unjustified wide use of MAR in statistics, and there are
well-developed methods based on much weaker assumptions \cite{manski2003}.
\par
As in the previous section, we want to refrain from making strong assumptions
about the mechanism that is behind the generation of missing values, apart
from what little is already implicit in the definition of the multi-valued map
$\mvm$. We have argued before that the information in $\mvm$, i.e., about the
relationship between $\srv$ and $\orv$, can be represented by the following
conditional lower prevision $\lpr(\cdot\vert\srv)$ on $\gambles(\statobs)$:
\begin{equation}
\label{eq:md-missingness}
\lpr(h\vert x)=\min_{o\in\mvm(x)}h(x,o),
\end{equation}
for all gambles $h$ on $\statobs$ and all $x\in\states$.
\par
We make the following additional \emph{irrelevance assumption}: for all
gambles $f$ on $\classes$,
\begin{equation}
\lpr(f\vert x,o)=\lpr_0(f\vert x)
\text{ for all  $x\in\states$ and $o\in\mvm(x)$}.
\tag{MDI}
\label{eq:md-irrelevance}
\end{equation}
Assumption~\eqref{eq:md-irrelevance} states that, conditional on the
attributes variable $\srv$, the observations variable $\orv$ is irrelevant to
the class, or in other words that the incomplete observations $o\in\mvm(x)$
can influence our beliefs about the class only indirectly through the value
$x$ of the attributes variable $X$. We shall discuss this assumption in more
detail at the end of this section.
\par
Summarising, we now have a coherent lower prevision $\lpr_0$ on
$\gambles(\states)$, a separately coherent conditional lower prevision
$\lpr(\cdot\vert\srv)$ on $\gambles(\states\times\observs)$, and a separately
coherent conditional lower prevision $\lpr(\cdot\vert\srv,\orv)$ on
$\gambles(\classes\times\states\times\observs)$, determined from
$\lpr_0(\cdot\vert\srv)$ through the irrelevance
assumption~\eqref{eq:md-irrelevance}.\footnote{Actually, the irrelevance
  assumption~\eqref{eq:md-irrelevance} does not determine
  $\lpr(\cdot\vert\srv,\orv)$ completely, but we shall see that this is of no
  consequence for finding the posterior $\rexl(\cdot\vert\orv)$.}  We can now
apply a generalisation of Walley's Marginal Extension Theorem (see
Theorem~\ref{theo:ext-mar-ext} in Appendix~\ref{sec:ext-mar-ext}), to find
that the smallest coherent lower prevision $\lpr$ on
$\gambles(\classes\times\states\times\observs)$ that has marginal $\lpr_0$ and
is jointly coherent with $\lpr(\cdot\vert\srv)$ and
$\lpr(\cdot\vert\srv,\orv)$, is given by
\begin{equation}
\label{eq:md-joint}
\lpr(h)=\lpr_0(\lpr(\lpr(h\vert\srv,\orv)\vert\srv)),
\end{equation}
for all gambles $h$ on $\classes\times\statobs$.
\par
We can now use regular extension to obtain the conditional lower prevision
$\rexl(\cdot\vert\orv)$ on $\gambles(\classes)$.  It yields the smallest (most
conservative) posterior lower prevision that is jointly coherent with $\lpr$
(and therefore with $\lpr_0$, $\lpr(\cdot\vert\states)$ and
$\lpr(\cdot\vert\states\times\observs)$) and satisfies an additional
regularity condition. Here too, it leads to the right way to obtain a
posterior lower prevision on the Bayesian sensitivity analysis interpretation.
Again, observe that using natural rather than regular extension would lead to
a \emph{completely} vacuous posterior on $\classes$.

\begin{theorem}[Conservative updating rule]
\label{theo:md-marco}
Assume that the irrelevance assumption~\eqref{eq:md-irrelevance} holds.  Let
$o$ be any element of\/ $\observs$. Then $\{o\}_*=\emptyset$. If
$\upr_0(\{x\})>0$ for all $x\in\{o\}^*$, then for any gamble $f$ on
$\classes$:
\begin{equation}
\label{eq:md-marco}
\rexl(f\vert o)=\min_{x\colon o\in\mvm(x)}\lpr_0(f\vert x).
\end{equation}
\end{theorem}

\begin{proof}
  Consider any $x=(a_1,\dots,a_n)$ in $\states$. Since, by
  Eq.~\eqref{eq:md-mvm}, $\mvm_k(a_k)=\{a_k,\ast\}$, we find that $\mvm(x)$
  can never be a singleton, whence indeed
\begin{equation*}
\{o\}_*=\set{x\in\states}{\mvm(x)=\{o\}}=\emptyset.
\end{equation*}
In order to calculate the regular extension $\rexl(f\vert o)$, the discussion
in Section~\ref{sec:regular-extension} tells us that we need to know the value
of $\upr(\classes\times\states\times\{o\})$.  Taking into account separate
coherence, we find that for all $(x,p)$ in $\statobs$,
\begin{equation*}
\upr(\classes\times\states\times\{o\}\vert x,p)
=\upr(I_{\classes\times\states\times\{o\}}(\cdot,x,p)\vert x,p)
=I_{\{o\}}(p)\lpr(\classes\vert x,p)=I_{\{o\}}(p),
\end{equation*}
whence $\upr(\classes\times\states\times\{o\}\vert\srv,\orv)=I_{\{o\}}$.
Consequently, we find for all $x\in\states$ that
\begin{equation*}
\upr(\upr(\classes\times\states\times\{o\}\vert\srv,\orv)\vert x)
=\max_{p\in\mvm(x)}I_{\{o\}}(p)
=\begin{cases}
1 & \text{if $o\in\mvm(x)$}\\
0 & \text{otherwise}
\end{cases}
=I_{\{o\}^*}(x),
\end{equation*}
whence $\upr(\upr(\classes\times\states\times\{o\}\vert\srv,\orv)\vert\srv)
=I_{\{o\}^*}$, and therefore, by Eq.~\eqref{eq:md-joint},
\begin{equation*}
\upr(\classes\times\states\times\{o\})
=\upr_0(\upr(\upr(\classes\times\states\times\{o\}
\vert\srv,\orv)\vert\srv))
=\upr_0(\{o\}^*)>0,
\end{equation*}
where the last inequality follows from the assumptions.  Since
$\upr(\classes\times\states\times\{o\})>0$, we can calculate the regular
extension as
\begin{equation*}
\rexl(f\vert o)
=\max\set{\mu}
{\lpr\left(I_{\classes\times\states\times\{o\}}[f-\mu]\right)\geq0}.
\end{equation*}
Again using separate coherence, we find that for all $(x,p)$ in $\statobs$,
\begin{multline*}
  \lpr(I_{\classes\times\states\times\{o\}}[f-\mu]\vert x,p)
  =\lpr(I_{\classes\times\states\times\{o\}}(\cdot,x,p)
  [f-\mu]\vert x,p)\\
  =I_{\{o\}}(p)\lpr(f-\mu\vert x,p) =I_{\{o\}}(p)[\lpr(f\vert x,p)-\mu],
\end{multline*}
whence $\lpr(I_{\classes\times\states\times\{o\}}[f-\mu]\vert\srv,\orv)
=I_{\{o\}}[\lpr(f\vert\srv,\orv)-\mu]$. Consequently, we find that for all
$x\in\states$, using Eq.~\eqref{eq:md-missingness} and the irrelevance
assumption~\eqref{eq:md-irrelevance},
\begin{multline*}
  \lpr(\lpr(I_{\classes\times\states\times\{o\}}[f-\mu]
  \vert\srv,\orv)\vert x)\\
\begin{aligned}
  &=\min_{p\in\mvm(x)}I_{\{o\}}(p)[\lpr(f\vert x,p)-\mu]
  =\min_{p\in\mvm(x)}I_{\{o\}}(p)[\lpr_0(f\vert x)-\mu]\\
  &=\begin{cases}
    \min\{0,\lpr_0(f\vert x)-\mu\} & \text{if $o\in\mvm(x)$}\\
    0 & \text{otherwise}
\end{cases}\\
&=I_{\{o\}^*}(x)\min\{\lpr_0(f\vert x)-\mu,0\},
\end{aligned}
\end{multline*}
where we used the fact that $\{o\}_*=\emptyset$.  Consequently,
$\lpr(\lpr(I_{\classes\times\states\times\{o\}}[f-\mu]
\vert\srv,\orv)\vert\srv)=I_{\{o\}^*}\min\{\lpr_0(f\vert\srv)-\mu,0\}$, and
therefore, by Eq.~\eqref{eq:md-joint},
\begin{multline*}
  \lpr(I_{\classes\times\states\times\{o\}}[f-\mu])
  =\lpr_0(\lpr(\lpr(I_{\classes\times\states\times\{o\}}[f-\mu]
  \vert\srv,\orv)\vert\srv))\\
  =\lpr_0(I_{\{o\}^*}\min\{\lpr_0(f\vert\srv)-\mu,0\}),
\end{multline*}
whence
\begin{multline*}
  \rexl(f\vert o) =\max\set{\mu}
  {\lpr\left(I_{\classes\times\states\times\{o\}}[f-\mu]\right)\geq0}\\
  =\max\set{\mu} {\lpr_0(I_{\{o\}^*}\min\{\lpr_0(f\vert\srv)-\mu,0\})\geq0}.
\end{multline*}
A course of reasoning similar to the one in the proof of
Theorem~\ref{theo:marco-simple} now tells us that indeed
\begin{equation*}
\rexl(f\vert o)=\min_{x\in\{o\}^*}\lpr_0(f\vert x)
\end{equation*}
[replace the gamble $f$ on $\states$ in that proof by the gamble
$\lpr_0(f\vert\srv)$].
\end{proof}

\subsection*{The conservative updating rule}
Let us now denote by $E$ that part of the attributes variable $\srv$ that is
instantiated, for which actual values are available. We denote its value by
$e$. Let $R$ denote the other part, for whose components values are missing.
We shall denote the set of its possible values by $\missings$, and a generic
element of that set by $r$. Observe that for every $r\in\missings$, the
attributes vector $(e,r)$ is a possible \emph{completion} of the incomplete
observation $o=(e,\ast)$ (with some abuse of notation) to a complete
attributes vector. Moreover, $\{o\}^*=\{e\}\times\missings$. We deduce from
Theorem~\ref{theo:md-marco} that the updated lower prevision $\rexl(\cdot\vert
e,\ast)$ is then given by
\begin{equation}
\label{eq:magic-formula}
\rexl(f\vert e,\ast)=\min_{r\in\missings}\lpr_0(f\vert e,r)
\tag{CUR}
\end{equation}
for all gambles $f$ on $\classes$, provided that $\upr_0(\{(e,r)\})>0$ for all
$r\in\missings$, which we shall assume to be the case. We shall
call~\eqref{eq:magic-formula} the \emph{conservative updating rule}.
\par
We shall discuss the case that $\lpr_0$ and $\lpr_0(\cdot\vert\srv)$ are
imprecise in Section~\ref{sec:credal nets}. But let us first, for the
remainder of this section, and in Sections~\ref{sec:bayesian-nets}
and~\ref{sec:algorithm}, assume that $\lpr_0$ and $\lpr_0(\cdot\vert\srv)$ are
precise.  Observe that even in this case, the posterior $\rexl(\cdot\vert
e,\ast)$ is imprecise.  How can we use this imprecise posterior to perform
classification? We shall only discuss the simplest case: we associate a reward
function $f_c=I_{\{c\}}$ with each class $c$ in $\classes$, and we look for
those classes $c$ that are undominated elements of the strict partial order
$\spref$ on $\classes$, defined by
\begin{align}
  c'\spref c''
  &\Leftrightarrow\rexl(I_{\{c'\}}-I_{\{c''\}}\vert e,*)>0\notag\\
  &\Leftrightarrow\min_{r\in\missings}
  \pr_0(I_{\{c'\}}-I_{\{c''\}}\vert e,r)>0\notag\\
  &\Leftrightarrow(\forall r\in\missings) (p_0(c'\vert e,r)>p_0(c''\vert e,r))
\label{eq:robustness}\\
&\Leftrightarrow\min_{r\in\missings} \dfrac{p_0(c'\vert e,r)}{p_0(c''\vert
  e,r)}>1,\notag
\end{align}
where we have used~\eqref{eq:magic-formula}, and where $p_0(\cdot\vert e,r)$
denotes the mass function of $\pr_0(\cdot\vert e,r)$. Since for all $r$ in
$\missings$, it is also assumed that $p_0(e,r)>0$, we can apply Bayes' rule to
rewrite this as
\begin{equation}
\label{eq:credal-dominance}
c'\spref c''\Leftrightarrow
\min_{r\in\missings}
\dfrac{p_0(c',e,r)}{p_0(c'',e,r)}>1.
\end{equation}
Eq.~\eqref{eq:robustness} is interesting: it tells us that $c'\spref c''$ if
$c'$ is strictly preferred to $c''$ under all the possible completions $(e,r)$
of the observed data $(e,\ast)$, i.e., if the strict preference is
\emph{robust} under all these possible completions.
\par
Classification is then done by assigning an object with observed attributes
$(e,\ast)$ to the \emph{set} of optimal, undominated classes for the strict
preference $\spref$. Among these optimal classes, there may be classes $c'$
and $c''$ that are equivalent:
\begin{equation*}
(\forall r\in\missings)
(p_0(c'\vert e,r)=p_0(c''\vert e,r)),
\end{equation*}
i.e., that are equally probable under all possible completions $(e,r)$ of
$(e,\ast)$. Otherwise they are incomparable, which means that $p_0(c'\vert
e,r_1)\geq p_0(c''\vert e,r_1)$ for some completion $(e,r_1)$ and $p_0(c'\vert
e,r_2)\leq p_0(c''\vert e,r_2)$ for another completion $(e,r_2)$, where one of
these inequalities will be strict.  For such incomparable classes, the fact
that observations are missing is responsible for our inability to make a
choice between them.
\par
In the case of the earlier medical example, $e$ denotes the part of the
profile that is known for a patient and the same incomplete profile can be
regarded as the set $\{(e,r)\vert r\in\missings\}$ of complete profiles that
are consistent with it. The conservative updating rule tells us that in order
to update our beliefs on the possible diseases given the incomplete profile,
we have to consider all the complete profiles consistent with it, which leads
us to lower and upper probabilities and previsions. As we explained above,
this will generally give rise only to partial classifications. That is, in
general we shall only be able to exclude some of the possible diseases given
the evidence. This \emph{may} lead to the identification of a single disease,
but only when the conditions justify precision.
\par
The conservative updating rule is a significant result: it provides us with
the correct updating rule to use with an unknown incompleteness mechanism; and
it shows that robust, conservative inference can be achieved by relying only
on the original prior model of domain uncertainty.
\par
It also is a conceptually simple rule, as it involves taking all the possible
completions of the missing attributes. It is not, therefore, very surprising
that the use of analogous procedures has already been advocated in the context
of robust statistical inference (see for instance
\cite{manski2003,ramoni2001b,zaffalon2002b}).  These focus on the problem of
\emph{learning} a model from an incomplete sample, which is then simply
regarded as the set of all the complete samples that are consistent with it.
But we are not aware of anyone proposing (and justifying) the same intuitive
principle for updating beliefs when observations are incomplete. Perhaps the reluctance to change firmly entrenched beliefs about the
more traditional naive updating has played a role in this. In contradistinction with the previous work on learning models, we are indeed
proposing a new (coherent) rule for \emph{updating beliefs}.

\subsection*{Some comments on the irrelevance assumption}
Let us end this section with a discussion of the irrelevance
assumption~\eqref{eq:md-irrelevance}, but placed in a context more general
than classification. [Additional technical comments on
Assumption~\eqref{eq:md-irrelevance} in the case that $\lpr_0$ and
$\lpr_0(\cdot\vert\srv)$ are precise, are given in
Appendix~\ref{sec:extra-independence}.]
\par
Assume that we are studying the relation between \emph{observations} $\srv$
and \emph{conclusions} $\crv$, in the sense that observing the value $x$ of
$\srv$ in $\states$ changes our beliefs about which value $\crv$ assumes in
$\classes$. Due to some reason, we cannot observe the value of $X$, but there
is an incompleteness mechanism that produces an incomplete version $\orv$ of
$\srv$. In this general context, Assumption~\eqref{eq:md-irrelevance} tells us
that if we have a precise observation $\srv=x$, then the additional knowledge
of what incomplete observation $\orv=o$ is generated by $x$, will not affect
our beliefs about the conclusion $\crv$. In other words, if we know the value
of the precise observation, then knowing what incomplete observation it
produces, becomes completely superfluous. This can be easily reformulated in
the more specific context of classification discussed above: if we know the
value of all the attributes, then knowing that some of the attributes fail to
be measured will be irrelevant to the classification.
\par
We feel that this is precisely what characterises problems of missing data, or
of incomplete observations: when something that can be missing is actually
measured, the problem of missing data disappears. Let us consider the opposite
case, where the bare fact that an attribute is not measured is directly
relevant to predicting the class. This fact should then become part of the
classification model by making a new attribute out of it, and treating it
accordingly, so that this should not be regarded as a problem of missing
information. Stated differently, once the model properly includes all the
factors that are relevant to predicting the class, \eqref{eq:md-irrelevance}
follows naturally.
\par
Regarding the relationship between assumption CAR/MAR and our irrelevance assumption~\eqref{eq:md-irrelevance}, it is not difficult to prove that if the former is satisfied (even in the case of an imprecise prior discussed in Theorem~\ref{theo:car}) then
the latter holds automatically.
This is not surprising as the CAR/MAR assumption identifies a subset of a much
larger class of incomplete observation (and missing data) problems, which are
characterised in general by~\eqref{eq:md-irrelevance}. Note, however, that
although one implies the other, they do refer to different things. In the
context of classification, MAR states that any incomplete observation $o$ is
equally likely to have been produced by all the attribute vectors $x$ that may
produce it, i.e., there is no compatible attribute vector $x$ that yields
observation $o$ with a higher probability $p(o\vert x)$ than any other
compatible attribute vector. MAR therefore says something about the mechanism
that produces observations $o$ from attribute vectors $x$, i.e., about the
\emph{the missing data mechanism} itself. Our irrelevance
condition~\eqref{eq:md-irrelevance}, on the other hand, states that if we know
the attribute vector precisely, then knowing in addition what observation $o$
is produced will not affect the classification.  In other words, we assume
that the classification only depends on the attributes, and \emph{not on the
  missing data mechanism}.
\par
CAR/MAR is much stronger than our irrelevance assumption, but it is worth
pointing out that there are cases where making the MAR assumption is
completely justified, and where, consequently, our approach leads to results
that are much too weak. We give one notable example: the case of an attribute
that we know is always missing.  In this case the missing data mechanism
clearly satisfies the MAR assumption: the probability of outcome $\ast$ is
one, irrespective of the actual value of the attribute. MAR then tells us that
we can discard this attribute variable, or `marginalise it out', as is the
usual practice. We should therefore not apply the conservative updating rule.
We advocate using our rule only when nothing is known about the incompleteness
mechanism, and this clearly is not the case here.
\par
It may useful to extend the discussion to statistical inference, even if,
strictly speaking, this goes beyond the scope of our present work.  In
particular, it is well-known (see for instance
\cite[Proposition~2.1]{manski2003}) that the CAR/MAR assumption cannot be
tested statistically, in the sense that we cannot use incomplete observations
to check whether it is reasonable. It does not seem to be possible to test
Assumption~\eqref{eq:md-irrelevance} either, for essentially the same reasons.
To understand this, let us, for the sake of simplicity, look at the case of
precise probabilities: it should be tested whether or not $p(c\vert
x,o)=p(c\vert x)$ for all classes $c$ (with obvious notations). The problem is
that the precise observation $x$ is always hidden to us; we can only see the
incomplete observation $o$. So in a statistical inference setting only
$p(c,o)$ and not $p(c,x,o)$ would be accessible via the data, and we would not
be able to perform the test. Therefore, there appears to exist a fundamental
limitation of statistical inference in the presence of missing data: the
actually observed data seem not to allow us to test our assumptions about the
missing data mechanism, but nevertheless our inferences rely heavily on the
specific assumptions that we make about it!  This is one of the reasons why we
are advocating that only those assumptions should be imposed that are weak
enough to be tenable. On our view, \eqref{eq:md-irrelevance} is a good
candidate.

\section{Classification in expert systems with Bayesian networks}
\label{sec:bayesian-nets} 
One popular way of doing classification in complex real-world domains involves
using \emph{Bayesian networks}.  These are precise probabilistic models
defined by a directed acyclic graph and a collection of conditional mass
functions \cite{pearl1988}.
\par
A generic node $Z$ in the graph is identified with a random variable taking
values in a finite set $\mathcal{Z}$ (we use `node' and `variable'
interchangeably, and we reserve the same symbol for both).  Each variable $Z$
holds a collection of conditional mass functions $p_0^{Z\vert\pi_Z}$, one for
each possible joint value $\pi_Z$ of its direct predecessor nodes (or
\emph{parents}) $\Pi_Z$. The generic conditional mass function
$p_0^{Z\vert\pi_Z}$ assigns the probability
$P_0(\{z\}\vert\pi_Z)=p_0(z\vert\pi_Z)$ to a value $z\in\mathcal{Z}$ (we drop
the superscript when we refer to actual probabilities.)

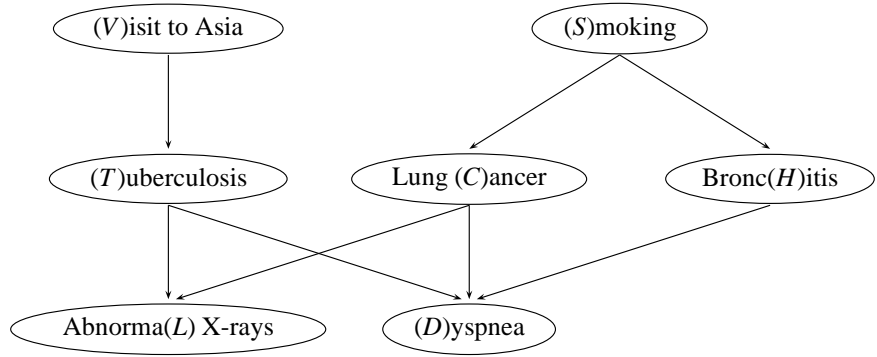
\begin{figure}
\psset{xunit=1mm,yunit=1mm,runit=1mm}
\begin{pspicture}(0,0)(100.00,50.00)
\rput(20.00,10.00){\psovalbox[linewidth=0.15,linecolor=black]{Abnorma($L$) X-rays}}
\rput(60.00,10.00){\psovalbox[linewidth=0.15,linecolor=black]{($D$)yspnea}}
\rput(20.00,30.00){\psovalbox[linewidth=0.15,linecolor=black]{($T$)uberculosis}}
\rput(20.00,50.00){\psovalbox[linewidth=0.15,linecolor=black]{($V$)isit to Asia}}
\rput(60.00,30.00){\psovalbox[linewidth=0.15,linecolor=black]{Lung ($C$)ancer}}
\rput(100.00,30.00){\psovalbox[linewidth=0.15,linecolor=black]{Bronc($H$)itis}}
\rput(80.00,50.00){\psovalbox[linewidth=0.15,linecolor=black]{($S$)moking}}
\psline[linewidth=0.15,linecolor=black]{->}(20.00,46.50)(20.00,34.00)
\psline[linewidth=0.15,linecolor=black]{->}(60.00,26.50)(60.00,14.00)
\psline[linewidth=0.15,linecolor=black]{->}(20.00,26.50)(20.00,14.00)
\psline[linewidth=0.15,linecolor=black]{->}(80.00,46.50)(60.00,34.00)
\psline[linewidth=0.15,linecolor=black]{->}(80.00,46.50)(100.00,34.00)
\psline[linewidth=0.15,linecolor=black]{->}(60.00,26.50)(21.00,14.00)
\psline[linewidth=0.15,linecolor=black]{->}(20.00,26.50)(59.00,14.00)
\psline[linewidth=0.15,linecolor=black]{->}(100.00,26.50)(61.00,14.00)
\end{pspicture}\\
\caption{The `Asia' Bayesian network.}
\label{fig:asia}
\end{figure}

Figure~\ref{fig:asia} displays the well-known example of Bayesian network
called `Asia'.\footnote{The network presented here is equivalent to the
  traditional one, although it is missing a logical OR node.} This models an
artificial medical problem by means of cause-effect relationships between
random variables, e.g., $S\to C$ (each variable is denoted for short by the
related letter between parentheses in Figure~\ref{fig:asia}). The variables
are binary and for any given variable, for instance $V$, its two possible
values are denoted by $v'$ and $v''$, for the values `yes' and `no',
respectively.  The conditional probabilities for the variables of the model
are reported in Table~\ref{tab:asia}.

\begin{table}[htbp]
\centering
\begin{tabular}{l|llllllll}
\hline \\
$V=v'$ & $0.01$ &  &  &  &  &  &  &  \\ \hline \\
$S=s'$ & $0.5$ &  &  &  &  &  &  &  \\ \hline \\
& $v'$ & $v''$ &  &  &  &  &  &  \\
$T=t'$ & $0.05$ & $0.01$ &  &  &  &  &  &  \\ \hline \\
& $s'$ & $s''$ &  &  &  &  &  &  \\
$C=c'$ & $0.1$ & $0.01$ &  &  &  &  &  &  \\ \hline \\
& $s'$ & $s''$ &  &  &  &  &  &  \\
$H=h'$ & $0.6$ & $0.3$ &  &  &  &  &  &  \\ \hline \\
& $t'c'$ & $t'c''$ & $t''c'$ & $t''c''$ &  &
&  &  \\
$L=l'$ & $0.98$ & $0.98$ & $0.98$ & $0.05$ &  &  &  &  \\ \hline \\
& $t'c'h'$ & $t'c'h''$ & $t'c''h'$ & $t'c''h''$
& $t''c'h'$ & $t''c'h''$ & $t''c''h'$ & $t''c''h''$ \\
$D=d'$ & $0.9$ & $0.7$ & $0.9$ & $0.7$ & $0.9$ & $0.7$ & $0.8$ & $0.1$ \\
\hline
\end{tabular}
\vspace{.5\baselineskip}
\caption{Asia example: probabilities for each variable (first column)
  in the graph conditional on the values of the parent variables
\label{tab:asia}}
\end{table}

Bayesian nets satisfy the \emph{Markov condition}: every variable is
stochastically independent of its non-descendant non-parents given its
parents. Let us consider a generic Bayesian network with nodes $C$, $A_1$,
$\ldots$, $A_n$ (for consistency with the notation in
Section~\ref{sec:missing-data}). From the Markov condition, it follows that
the joint mass function $p_0$ is given by
\begin{equation}
p_0(c,a_1,\ldots ,a_n)
=p_0(c\vert \pi_C)\prod_{i=1}^{n}p_0(a_{i}\vert\pi_{A_i})\quad
\forall(c,a_1,\cdots ,a_n)\in\mathcal{C}\times\mathcal{X},
\label{factorisation}
\end{equation}
where the values of the parent variables are those consistent with
$(c,a_1\ldots ,a_n)$. Hence, a Bayesian network is equivalent to a joint mass
function over the variables of the graph. We assume that such a joint mass
function assigns positive probability to any event.
\par
Bayesian nets play an important role in the design of expert systems.  In this
case, domain experts are supposed to provide both the qualitative graphical
structure and the numerical values for the probabilities, thus implicitly
defining an overall model of the prior uncertainty for the domain of interest.
Users can then query the expert system for updating the marginal prior
probability of $C$ to a posterior probability according to the available
evidence $E=e$, i.e., a set of nodes with known values. In the Asia net, one
might ask for the updated probability of lung cancer ($C=c'$), given that a
patient is a smoker ($S=s'$) and has abnormal X-rays ($L=l'$), aiming
ultimately at making the proper diagnosis for the patient. This kind of
updating is very useful as it enables users to do classification, along the
lines given in Section~\ref{sec:missing-data}.

\subsection*{On updating probabilities with Bayesian networks}
Updating the uncertainty for the class variable in a Bayesian net is subject
to the considerations concerning incomplete observations in the preceding
sections, as generally the evidence set $E$ will not contain all the
attributes. To address this problem, one can assume that MAR holds and
correspondingly use the naive updating rule to get the posterior
$p_0(c\vert\{e\}\times\missings)$, but we have already pointed out that this
approach is likely to be problematical in real applications.  Nevertheless,
assuming MAR seems to be the most popular choice with Bayesian nets and the
literature presents plenty of algorithmic developments dealing with this case.
\par
Peot and Shachter \cite{peot1998} are a notable exception. In their paper,
they explicitly report that ``the current practice for modelling missing
observations in interactive Bayesian expert systems is incorrect.'' They show
this by focusing on the medical domain where there exists a systematic (i.e.,
non-MAR) incompleteness mechanism originated by the user of the expert system
and also by the patient himself. Indeed, there is a bias in reporting, and
asking for, symptoms that are present instead of symptoms that are absent; and
a bias to report, and ask for, urgent symptoms over the others. Peot and
Shachter tackle this problem by proposing a model of the incompleteness
mechanism for the specific situation under study. Explicitly modelling the
missing data mechanism is in fact another way to cope with the problem of
incomplete observations, perhaps involving the same Bayesian net. The net
would then also comprise the nodes $\orv_k$, $k=1,\dots ,n$, for the
incomplete observations; and the posterior probability of interest would
become $p(c\vert o)$. Unfortunately, this approach presents serious practical
difficulties.  Modelling the mechanism can be as complex as modelling the
prior uncertainty.  Furthermore, it can be argued that in contrast with domain
knowledge (e.g., medical knowledge), the way information can be accessed
depends on the particular environment where a system will be used; and this
means that models of the missing data mechanism will probably not be
re-usable, and therefore costly.
\par
These considerations support adopting a robust approach that can be
effectively implemented, like the one we proposed in
Section~\ref{sec:missing-data}. It is also useful to stress that our approach
has quite general applicability. The conservative updating rule, for example,
is perfectly suited to addressing Peot and Shachter's problem, as the biases
they deal with are easily shown to satisfy the irrelevance condition
\eqref{eq:md-irrelevance}.
\par
We next develop an algorithm that exploits~\eqref{eq:magic-formula} to perform
reliable classification with Bayesian networks.

\section{An algorithm to classify incomplete evidence with Bayesian networks}
\label{sec:algorithm} In this section we develop an algorithm to
perform classification with Bayesian networks by using the conservative
updating rule~\eqref{eq:magic-formula}. As discussed in
Section~\ref{sec:decision-making} and later at the end of
Section~\ref{sec:missing-data}, it is important to realise first that
conservative updating will not always allow two classes to be compared, i.e.,
\eqref{eq:magic-formula} generally produces only a partial order on the
classes.
\par
As a consequence, the classification procedure consists in comparing each pair
of classes by strict preference (which we shall also call \emph{credal
  dominance}, in accordance with \cite{zaffalon2000}) and in discarding the
dominated ones. The system will then output a set of \emph{possible}, optimal
classes. In the following we address the issue of efficient computation of the
credal dominance test. Let $c'$ and $c''$ be two classes in $\mathcal{C}$. We
shall use Eq.~\eqref{eq:credal-dominance} to test whether $c'$
credal-dominates $c''$.
\par
Let $\pi'$ and $\pi''$ denote values of the parent variables consistent with
the completions $(c',e,r)$ and $(c'',e,r)$, respectively. If a node's parents
do not contain $C$, let $\pi$ denote the value of the parent variables
consistent with $(e,r)$. With some abuse of notation, we shall treat the
vector $R$ of those attributes for which measurements are missing, in the
following as a set.  Furthermore, without loss of generality, let $A_1,\ldots
, A_m$, $m\leq n$, be the \emph{children} (i.e., the direct successor nodes)
of $C$, and $K=\{1,\ldots , m\}$. We shall denote $C$ in the following also as
$A_0$. For each $i=0,\ldots,m$, let
$\Pi^+_{A_i}=\Pi_{A_i}\cup\left\{A_i\right\}$.  Consider the functions
$\phi_{A_i}\colon\times_{j\colon A_j\in\Pi_{A_i}^+\cap
  R}\mathcal{A}_j\to\mathbb{R}^+$ ($i=0,\dots,m$), with values equal
to ${p_0(a_i\vert \pi_{A_i}')}/{p_0(a_i\vert\pi_{A_i}'')}$ for $i\in K$, and
equal to ${p_0(c'\vert\pi_C)}/{p_0(c''\vert\pi_C)}$ for $i=0$. We use the
symbol $\mu$ to denote the minima of the $\phi$-functions, in the following
way:
\begin{align}
\mu_{A_0}
&=\min_{\substack{a_j\in\mathcal{A}_j,\\A_j\in\Pi_C^+\cap R}}
\frac{p_0(c'\vert\pi_C)}{p_0(c''\vert\pi_C)}\\
\mu_{A_i}
&=\min_{\substack{a_j\in\mathcal{A}_j,\\A_j\in\Pi_{A_i}^+\cap R}}
\frac{p_0(a_i\vert \pi_{A_i}')}{p_0(a_i\vert\pi_{A_i}'')}, 
\quad i\in K.
\end{align}
Consider the \emph{Markov blanket} of $C$, that is, the set of nodes
consisting of the parents of $C$, its children, and the parents of the
children of $C$. Denote by $B^+$ the union of $C$ with its Markov blanket. We
shall refer to $B^+$ both as a set of nodes and as a subgraph, depending on
the context. Initially we focus on networks for which $B^+$ is singly
connected (the overall network can still be multiply connected). We have the
following result.

\begin{theorem}
\label{theo:algorithm} 
Consider a Bayesian network with nodes $C$, $A_1$, \dots, $A_n$, for which
$B^+$ is singly connected. Let $c',c''\in\mathcal{C}$. Then $c'$
credal-dominates $c''$ if and only if\/ $\prod_{i=0}^{m}\mu_{A_i}>1$.
\end{theorem}

\begin{proof}
 Rewrite the minimum in Eq.~\eqref{eq:credal-dominance} as follows:
\begin{align}
\min_{r\in\missings}\frac{p_0(c',e,r)}{p_0(c'',e,r)} 
&=\min_{r\in\missings}
\left[\frac{p_0(c'\vert\pi_C)}{p_0(c''\vert\pi_C)} 
\prod_{i\in K}\frac{p_0(a_i\vert\pi_{A_{i}}')}{p_0(a_i\vert\pi_{A_{i}}'')}
\prod_{j\notin K}\frac{p_0(a_j\vert\pi_{A_{j}})}{p_0(a_j\vert\pi_{A_{j}})}
\right]\notag\\
&=\min_{\substack{a_j\in\mathcal{A}_j,\\ A_j\in B^+\cap R}}
\left[\frac{p_0(c'\vert\pi_C)}{p_0(c''\vert\pi_C)} 
\prod_{i\in K}\frac{p_0(a_i\vert\pi_{A_{i}}')}{p_0(a_i\vert\pi_{A_{i}}'')}
\right]\label{eq:cd2}.
\end{align}
This shows that the variables that do not belong to $B^+$ can be discarded in
order to test credal dominance. Now recall that every function $\phi_{A_i}$
[that is, every ratio in Eq.~\eqref{eq:cd2}] depends only on the variables in
$\Pi_{A_i}^+\cap R$. Given that $B^+$ is singly connected, we have that only
$\phi_{A_i}$ depends on the variables in $\Pi_{A_i}^+\cap R$. Let us show the
last statement by contradiction, by assuming that another function
$\phi_{A_k}$ ($k\in\{0,\ldots,m\}\setminus{\{i\}}$) depends on a variable in
$\Pi_{A_i}^+\cap R$. There are two cases, either the variable in
$\Pi_{A_i}^+\cap R$ is $A_i$ or it is a parent of $A_i$, say $U$.
\par
In the first case, neither $A_i$ nor $A_k$ coincide with the class variable
$C$: $A_i$ does not coincide with $C$ because no $\phi$-function depends on
$C$; in order for $\phi_{A_k}$ to depend on $A_i$, $A_i$ must be a parent of
$A_k$, so $A_i$ is not a child of $A_k$, whence $A_k$ cannot coincide with
$C$.  But $A_i$ being a parent of $A_k$ would create the undirected loop
$C$--$A_i$--$A_k$--$C$, making $B^+$ multiply connected. This case is
impossible.
\par
Consider now the second case when $\phi_{A_k}$ depends on $U$. In this case
$U$ must be a parent of $A_k$, besides being a parent of $A_i$. Note that $U$
does not coincide with $C$ because no $\phi$-function depends on $C$. As
before, these conditions imply that $B^+$ should be multiply connected. In the
case that $A_k$ coincides with $C$, the loop is $U$--$C$--$A_i$--$U$. If $C$
coincides with $A_i$, the loop is $U$--$C$--$A_k$--$U$. When neither $A_k$ nor
$A_i$ coincide with $C$, the loop is $U$--$A_k$--$C$--$A_i$--$U$.  In every
case we have a contradiction.
\par
Since the variables in $\Pi_{A_i}^+\cap R$ appear only in the argument of
$\phi_{A_i}$, they can be minimised out locally to $A_i$, obtaining
$\mu_{A_i}$. (Observe that $\mu_{A_i}$ is a number because only the variables
in $\Pi_{A_i}^+\cap R$ are in the argument of $\phi_{A_i}$.) Then the thesis
follows immediately.
\end{proof}

Theorem~\ref{theo:algorithm} renders the solution of the credal-dominance test
very easy when $B^+$ is singly connected,\footnote{This corrects the invalid
  claim, made in an earlier version of this paper \cite{cooman2003b}, that the
  complexity is linear for all networks.} with overall computational
complexity linear in the size of the input, i.e., $B^+$ (more precisely, the
input is the Bayesian network restricted to $B^+$). It is useful to emphasise
that the theorem works also for networks in which $B^+$ is multiply connected,
provided that the evidence $E=e$ makes $B^+$ become singly connected. Indeed
it is well known with Bayesian networks that the arcs leaving evidence nodes
can be removed while preserving the value $p_0(c\vert e)$ ($c\in\mathcal{C}$)
represented by the network. This result extends to credal dominance because it
is computed by $\min_{r\in\missings}[{p_0(c'\vert e,r)}/{p_0(c''\vert e,r)}]$
and because $p_0(c\vert e,r)$ is preserved by dropping the arcs leaving $E$,
for each $c\in\mathcal{C}$ and $r\in\mathcal{R}$.
\par
Now we move to the case that $B^+$ is multiply connected, and show how the
ideas behind the traditional way of dealing with multiply connected networks,
called \emph{conditioning}, can be applied here as well. Conditioning
\cite{pearl1988} works by instantiating a subset of nodes called the
\emph{loop cutset}. The removal of the arcs leaving the loop cutset creates a
singly connected net. The computation is then carried out on the singly
connected net as many times as there are joint states of the variables in the
cutset, and the results are eventually summarised to obtain the result related
to the multiply connected net.
\par
With credal dominance, the situation is analogous. We assume that the arcs
leaving evidence nodes in $B^+$ have been removed, and that a
loop cutset is given that opens the remaining loops (recall
that, according to the above observation, the loops are opened by the cutset
also where credal dominance is concerned). Call $R_1$ the loop cutset, and let
$R_2$ be the set of nodes such that $R=R_1\cup R_2$. Rewrite the test of
credal dominance as
\begin{equation*}
\min_{r_1\in\mathcal{R}_1}
\left[\min_{r_2\in\mathcal{R}_2}
\frac{p_0(c'\vert e,r_1,r_2)}{p_0(c''\vert e,r_1,r_2)}\right].
\end{equation*}
The inner minimisation is computed by Theorem~\ref{theo:algorithm} on the
graph $B^+$ made singly connected by dropping the arcs leaving $E\cup
R_1$. The outer minimisation is a simple enumeration of the states of the loop
cutset, which takes exponential time in general.
\par
From the viewpoint of worst-case computation complexity, the situation is
similar to the computation of the updating. However, the computation of credal
dominance will be easier in the cases where $B^+$ does not
coincide with the entire network. Furthermore, since $B^+$ can be singly
connected even when the network is multiply connected, the computation will be
linear also on some multiply connected nets.

\subsection*{An example}
\label{sec:example}
Let us consider the Asia net, where we choose $C$ as the class and set the the
evidence to $L=l'$ and $S=s'$. We want to test whether $c'$ credal-dominates
$c''$.
\par
Dropping the arcs leaving $S$, we obtain a new network in which $B^+$
is $\{C,L,D,T,H\}$. $B^+$ is multiply connected, and we select
$\{T\}$ as loop cutset. We start by considering the case $T=t'$. We must
compute $\mu_D$, $\mu_L$, and $\mu_C$. We have:
\begin{align*}
\mu_D
&=\min_{d\in\mathcal{D},h\in\mathcal{H}}
\frac{p_0(d\vert t',c',h)}{p_0(d\vert t',c'',h)}
=\min\left\{\frac{0.9}{0.9},\frac{0.7}{0.7},\frac{0.1}{0.1},\frac{0.3}{0.3}
\right\}=1\\
\mu_L
&=\frac{p_0(l'\vert t',c')}{p_0(l'\vert t',c'')}=\frac{0.98}{0.98}=1\\
\mu_C
&=\frac{p_0(c'\vert s')}{p_0(c''\vert s')}=\frac{0.1}{0.9}=\frac{1}{9},
\end{align*}
and their product is $1/9$. In the case $T=t''$, we obtain the following
values,
\begin{align*}
\mu_D
&=\min_{d\in\mathcal{D},h\in\mathcal{H}}
\frac{p_0(d\vert t'',c',h)}{p_0(d\vert t'',c'',h)}
=\min\left\{\frac{0.9}{0.8},\frac{0.7}{0.1},\frac{0.1}{0.2},\frac{0.3}{0.9}
\right\}=\frac{1}{3}\\
\mu_L
&=\frac{p_0(l'\vert t',c')}{p_0(l'\vert t',c'')}
=\frac{0.98}{0.05}=\frac{98}{5}\\
\mu_C
&=\frac{p_0(c'\vert s')}{p_0(c''\vert s')}
=\frac{0.1}{0.9}=\frac{1}{9},
\end{align*}
with product equal to $98/135\simeq 0.726$. The minimum of the products
obtained with the two values for $T$ is just $1/9$, so that $c''$ is
undominated.
\par
Testing whether $c''$ credal-dominates $c'$ is very similar and leads to
${45}/{686}$ as the value of the test, so $c'$ is undominated as well. In
this situation, the system suspends judgement, i.e., it outputs both the
classes, as there is not enough information to allow us to choose between the
two. This can be seen also by computing the posterior interval of probability for $c'$ by the
conservative updating rule, which leads to $[0.1,0.934]$. The width of this interval quantifies the 
mentioned lack of information. All of this should be contrasted with naive updating, which produces
$p_0(c'\vert l',s')\simeq 0.646$, and leads us to diagnose cancer.
\par
It is useful to better analyse the reasons for the indeterminate output of the
proposed system. Given our assumptions, the system cannot exclude that the
available evidence is part of a more complete piece of evidence where $T=t'$,
$D=d'$, and $H=h'$. If this were the case, then $c''$ would be nine times as
probable \emph{a posteriori} as $c'$, and we should diagnose no cancer.
However, the system cannot exclude either that the more complete evidence
would be $T=t''$, $D=d'$, and $H=h''$. In this case, the ratio of the
posterior probability of $c'$ to that of $c''$ would be ${686}/{45}$, leading
us to the opposite diagnosis.
\par
Of course when the evidence is strong enough, the proposed system does produce
determinate conclusions. For instance, the evidence $L=l'$, $S=s'$ and $T=t'$
will make the system exclude the presence of cancer.

\section{Working with credal networks}
\label{sec:credal nets}
Credal networks provide a convenient way of specifying prior knowledge using
the theory of coherent lower previsions.  They extend the formalism of
Bayesian networks by allowing sets of mass functions
\cite{cozman2000,fagiuoli1998}, or equivalently, sets of linear previsions.
These are also called \emph{credal sets} after Levi \cite{levi1980a}. We
recall that a credal set is equivalent to a coherent lower prevision, as
pointed out in Section~\ref{sec:credal sets}.
\par
A \emph{credal network} is a pair composed of a directed acyclic graph and a
collection of conditional credal sets\footnote{In this context, as in
  \cite{cozman2000}, we restrict ourselves to credal sets with a finite number
  of extreme points.} (i.e., a collection of conditional lower previsions). We
intend the graph to code strong independences. Two variables $Z_1$ and $Z_2$
are said to be \emph{strongly independent} when every vertex in the credal set
of joint mass functions for $(Z_1,Z_2)$, satisfies stochastic independence of
$Z_1$ and $Z_2$. That is, for every extreme mass function $p$ in the credal
set, and for all the possible pairs $(z_1,z_2)\in \mathcal{Z}_1\times
\mathcal{Z}_2$, it holds that $p(z_1|z_2)=p(z_1)$ and
$p(z_2|z_1)=p(z_2)$.\footnote{See also \cite{moral2002} for a complete account
  of different strong independence concepts and \cite{cozman2000a} for a deep
  analysis of strong independence.}  Each variable $Z$ in the net holds a
collection of conditional lower previsions, denoted by $\lpr_0^{Z\vert\pi_Z}$,
one for each possible joint value $\pi_Z$ of the node Z's parents $\Pi_Z$.
With some abuse of notation,\footnote{In preceding sections, the symbol
  $\solp$ was used to denote the dominating set of linear previsions. We use
  the same symbol here as there is one-to-one correspondence between linear
  previsions and mass functions (see Section~\ref{sec:credal sets}).} let
$\solp(\lpr_0^{Z\vert\pi_Z})$ be the credal set of mass functions for the
linear previsions dominating $\lpr_0^{Z\vert\pi_Z}$.
$p_0^{Z\vert\pi_Z}\in\solp(\lpr_0^{Z\vert\pi_Z})$ assigns the probability
$p_0(z\vert\pi_Z)$ to a value $z\in\mathcal{Z}$. In the following we assume
that each of these mass functions assigns positive probability to any event.
Given the equivalence between lower probability functions and credal sets, we
can regard each node of the net to hold a collection of conditional, so-called
\emph{local}, credal sets. Actually, the usual approach of specifying the
conditional lower previsions for the nodes precisely amounts to providing the
local credal sets directly.  This is commonly done by \emph{separately
  specifying} these credal sets \cite{ferreira2002,walley1991}, something that
we also assume here: this implies that selecting a mass function from a credal
set does not influence the possible choices in others. This assumption is
natural within a Bayesian sensitivity analysis interpretation of credal nets.
\par
Credal nets satisfy a generalised version of the Markov condition called the
\emph{strong Markov condition}: each variable is strongly independent of its
non-descendant non-parents given its parents. This leads immediately to the
definition of the \emph{strong extension} \cite{cozman2000a} of a credal net.
This is the most conservative lower prevision $\lpr_0$ on
$\gambles(\classes\times\states)$ that coherently extends the nodes'
conditional lower previsions, subject to the strong Markov condition.  Let the
nodes of the network be $C$ (i.e., $A_0$), $A_1$, $\ldots$, $A_n$, as before.
It is well known that the credal set equivalent to $\lpr_0$ is
\begin{equation}
\solp(\lpr_0)
=\convhull
\left\{p_0\text{ factorising as in Eq.~\eqref{factorisation}}\colon
p_0^{A_{i}\vert\pi_{A_{i}}}\in\solp(\lpr_0^{A_{i}\vert\pi_{A_{i}}}),
i=0,\dots,n\right\},
\label{eq:strong-extension}
\end{equation}
where $\convhull$ denotes the convex hull operation. In other words,
$\solp(\lpr_0)$ is the convex hull of the set of all the joint mass functions
that factorise according to Eq.~\eqref{factorisation}, obtained by selecting
conditional mass functions from the local credal sets of the net in all the
possible ways. The strong extension is an imprecise prior defined by means of
the composition of local information. From yet another viewpoint, the credal
set $\solp(\lpr_0)$ makes a Bayesian sensitivity analysis interpretation of
credal nets very natural: working with a credal net can equivalently be
regarded as working simultaneously with the set of all Bayesian nets
consistent with $\solp(\lpr_0)$.
\par
The credal set $\solp(\lpr_0)$ can have a huge number of extreme mass
functions. Indeed, the computation of lower and upper probabilities with
strong extensions is NP-hard \cite{ferreira2002}\footnote{However, it should
  be observed that Ferreira da Rocha and Cozman's result is proved for the
  subset of polytrees in which the local credal sets are convex hulls of
  degenerate mass functions that assign all the mass to one elementary event.
  As such, it does not tell us anything about the complexity of working with
  the case of polytrees whose credal sets are made up of mass
  functions that assign positive probability to any event.} also when the
graph is a polytree. Polytrees are directed acyclic graphs with the
characteristic that forgetting the direction of arcs, the resulting graph has
no undirected cycles. This should be contrasted with Bayesian networks for
which common computations take polynomial time with polytrees. Indeed, the
difficulty of computation with credal nets has severely limited their use so
far, even though credal nets have the great advantage over Bayesian nets of
not requiring the model probabilities to be specified precisely. This is a key
point for faithfully modelling human knowledge, which also allows expert
systems to be developed quickly.
\par
In the following we extend Theorem~\ref{theo:algorithm} to credal nets,
showing that conservative updating allows classification with credal nets to
be realised with the same complexity needed for Bayesian nets.  This
appears to be an important result, with implications for the practical
usability of credal nets in modelling knowledge.
\par
Below we reuse the definition $\Pi^+_{A_i}$ given in
Section~\ref{sec:algorithm}, we again denote by $B^+$ the union of $C$ with
its Markov blanket, and we refer to $C$ also by $A_{0}$. Consider the
following quantities:
\begin{equation}
p_{0\ast}^{C\vert\pi_{C}}
=\argmin_{p_0^{C\vert\pi_{C}}\in\solp(\lpr_0^{C\vert\pi_{C}})}
{\frac{p_0(c'\vert\pi_C)}{p_0(c''\vert\pi_C)}},
\label{past}
\end{equation}
and, for each $i\in K$,
\begin{align}
  \lms_0(a_i\vert \pi_{A_i}') &=\min_{p_0^{A_{i}\vert\pi_{A_{i}}'}
    \in\solp(\lpr_0^{A_{i}\vert\pi_{A_{i}}'})}p_0(a_i\vert \pi_{A_i}')
\label{lms}\\
\ums_0(a_i\vert \pi_{A_i}'') &=\max_{p_0^{A_{i}\vert\pi_{A_{i}}''}
  \in\solp(\lpr_0^{A_{i}\vert\pi_{A_{i}}''})}p_0(a_i\vert \pi_{A_i}''),
\label{ums}
\end{align}
as well as the functions
$\underline{\phi}_{A_i}\colon\times_{j\colon A_j\in\Pi_{A_i}^+\cap R}
\mathcal{A}_j\to\mathbb{R}^+$ ($i=0,\dots,m$), with values equal
to ${\lms(a_i\vert \pi_{A_i}')}/{\ums(a_i\vert\pi_{A_i}'')}$ for $i\in K$, and
equal to ${p_{0\ast}(c'\vert\pi_C)}/{p_{0\ast}(c''\vert\pi_C)}$ for $i=0$. We
use the symbol $\underline{\mu}$ to denote the minima of the
$\underline{\phi}$-functions, as follows:
\begin{align}
  \lmu_{A_0}&=\min_{\substack{a_j\in\mathcal{A}_j,\\
      A_j\in\Pi_C^+\cap R}}
  \frac{p_{0\ast}(c'\vert\pi_C)}{p_{0\ast}(c''\vert\pi_C)}\label{eq:ccd3.3}\\
  \lmu_{A_i}&=\min_{\substack{a_j\in\mathcal{A}_j\\
      A_j\in\Pi_{A_i}^+\cap R}} \frac{\lms_0(a_i\vert
    \pi_{A_i}')}{\ums_0(a_i\vert\pi_{A_i}'')},\quad i\in K.\label{eq:ccd3.2}
\end{align}
We have the following result.

\begin{theorem}
\label{theo:credal-algorithm}
Consider a credal net with nodes $C$, $A_1$, \dots, $A_n$, for which $B^+$ is
singly connected. Let $c',c''\in\mathcal{C}$. Then $c'$ credal-dominates $c''$
if and only if\/ $\prod_{i=0}^m\lmu_{A_i}>1$.
\end{theorem}

\begin{proof}
\label{}
A credal net can equivalently be regarded as a set of Bayesian nets, as is
apparent from Eq.~\eqref{eq:strong-extension}. Accordingly, for credal
dominance to hold with a credal net, it is necessary that it holds for all the
joint mass functions consistent with the strong extension.  This can be tested
by solving the following double minimisation problem:
\begin{align}
&\min_{p_0\in\solp(\lpr_0)}\min_{r\in\missings}
\frac{p_0(c',e,r)}{p_0(c'',e,r)}\label{ccdth2}\\
=&\min_{p_0^{C\vert\pi_{C}}\in\solp(\lpr_0^{C\vert\pi_{C}})}
\min_{\substack{p_0^{A_{k}\vert\pi_{A_{k}}'}
\in\solp(\lpr_0^{A_{k}\vert\pi_{A_{k}}'}),\\
p_0^{A_{k}\vert\pi_{A_{k}}''}\in\solp(\lpr_0^{A_{k}\vert\pi_{A_{k}}''}),\\
k\in K}} \min_{\substack{a_j\in\mathcal{A}_j,\\ A_j\in B^+\cap R}}
\left[\frac{p_0(c'\vert\pi_C)}{p_0(c''\vert\pi_C)} \prod_{i\in K}
\frac{p_0(a_i\vert\pi_{A_{i}}')}{p_0(a_i\vert\pi_{A_{i}}'')} \right]
\label{ccd2th2}\\
=&\min_{\substack{a_j\in\mathcal{A}_j,\\ A_j\in B^+\cap R}} \left\{
  \min_{p_0^{C\vert\pi_{C}}\in\solp(\lpr_0^{C\vert\pi_{C}})}
  \left[\frac{p_0(c'\vert\pi_C)}{p_0(c''\vert\pi_C)}\right] \prod_{i\in K}
  \frac{\min_{p_0^{A_{i}\vert\pi_{A_{i}}'}
      \in\solp(\lpr_0^{A_{i}\vert\pi_{A_{i}}'})}p_0(a_i\vert\pi_{A_{i}}')}
  {\max_{p_0^{A_{i}\vert\pi_{A_{i}}''}
      \in\solp(\lpr_0^{A_{i}\vert\pi_{A_{i}}''})}p_0(a_i\vert\pi_{A_{i}}'')}
\right\}\label{ccd3th2}\\
=&\min_{\substack{a_j\in\mathcal{A}_j,\\ A_j\in B^+\cap R}} \left[
  \frac{p_{0\ast}(c'\vert\pi_C)}{p_{0\ast}(c''\vert\pi_C)} \prod_{i\in K}
  \frac{\lms(a_i\vert\pi_{A_{i}}')}{\ums(a_i\vert\pi_{A_{i}}'')}
\right],
\label{ccd4th2}
\end{align}
where the passage from \eqref{ccdth2} to \eqref{ccd2th2} is due to
\eqref{eq:cd2} and \eqref{eq:strong-extension};\footnote{Actually, the passage
  is also based on the fact that the minimum of \eqref{ccdth2} is achieved at
  an extreme point of $\solp(\lpr_0)$. This is well-known with credal networks
  and is pointed out formally by Theorems 5 and 7 in reference
  \cite{fagiuoli1998}.} and the following passage is possible thanks to the
characteristic of separate specification of credal sets in the credal network.
Note that Expression~\eqref{ccd4th2} resembles Expression~\eqref{eq:cd2} of
Theorem~\ref{theo:algorithm}. In fact, the proof of
Theorem~\ref{theo:algorithm} below Expression~\eqref{ccd3th2} applies here as
well: $\underline{\phi}_{A_i}$ depends only on the variables in
$\Pi_{A_i}^+\cap R$ and only $\underline{\phi}_{A_i}$ depends on them. As in
Theorem~\ref{theo:algorithm}, the thesis follows immediately since the
variables in $\Pi_{A_i}^+\cap R$ can then be minimised out locally to $A_i$,
obtaining $\underline{\mu}_{A_i}$.
\end{proof}

Theorem~\ref{theo:credal-algorithm} renders the solution of the credal
dominance test for credal networks very easy when $B^+$ is singly connected.
However, in order to have a better idea of the computational complexity, one
has to carefully examine the complexity of solving
Problems~\eqref{past}--\eqref{ums}. This is what we set out to do in the
following.
\par
Let again $Z$ be a generic variable in the network. We consider three common
ways of specifying the local credal sets of the net.
\begin{enumerate}[1.]
\item In the first case, the conditional\footnote{The situation with root
    nodes is analogous.} credal set $\solp(\lpr_0^{Z|\pi_Z})$ for the variable
  $Z$ is specified via linear constraints on the probabilities $p_0(z|\pi_Z)$,
  $z\in\mathcal{Z}$. That is, in this representation the vector of
  probabilities $p_0(z|\pi_Z)$, $z\in\mathcal{Z}$, can take every value in a
  closed and bounded space described by linear constraints on the variables
  $p_0(z|\pi_Z)$, i.e., in a \emph{polytope}.\label{constraints}
\item In the second case, we assume that $\solp(\lpr_0^{Z|\pi_Z})$ is the
  convex hull of a set of mass functions directly provided by the
  modeller.\label{masses}
\item Finally, we consider the case when $\solp(\lpr_0^{Z|\pi_Z})$ is provided
  by specifying intervals of probability for the elementary events
  $(z|\pi_Z)$, $z\in\mathcal{Z}$. This is a special case of
  Case~\ref{constraints} where the only constraints allowed on the
  probabilities $p_0(z|\pi_Z)$ are bounds, except for
  $\sum_{z\in\mathcal{Z}}p_0(z|\pi_Z)=1$. Without loss of generality, we
  assume that the probability intervals are \emph{reachable}
  \cite{campos1994}. This holds if and only if $\solp(\lpr_0^{Z|\pi_Z})$ is
  nonempty and the intervals are tight, i.e., for each lower and upper bound
  there is a mass function in $\solp(\lpr_0^{Z|\pi_Z})$ at which the bound is
  attained. Reachable intervals produce a coherent lower prevision
  $\lpr_0^{Z|\pi_Z}$ that is \emph{2-monotone} \cite{campos1994}. For
  2-monotone lower previsions it holds that, given any two mutually exclusive
  events $\mathcal{Z}',\mathcal{Z}''\subseteq\mathcal{Z}$, there is a mass
  function $p_{0+}^{Z|\pi_Z}\in\solp(\lpr_0^{Z|\pi_Z})$ for which
  $\lms_0(\mathcal{Z}'|\pi_Z)=p_{0+}(\mathcal{Z}'|\pi_Z)$ and
  $\ums_0(\mathcal{Z}''|\pi_Z)=p_{0+}(\mathcal{Z}''|\pi_Z)$. We shall use this
  property in the following.\label{intervals}
\end{enumerate}
Observe that the representations in Cases~\ref{constraints} and~\ref{masses}
are fully general as any credal set can be represented by one or by the other.
In the following we consider that all the local credal set of the net are
specified either as in Case~\ref{constraints} or~\ref{masses}
or~\ref{intervals}. We do not consider mixed cases, which should be easy to
work out once the `pure' cases have been addressed.
\par
Let us now focus on the complexity of testing credal dominance in
Case~\ref{constraints}. Let $S$ be the size of the largest local credal set in
the network. The size is defined as the dimension of the constraints-variables
matrix that describes the linear domain. Let $O(L(S))$ be the complexity to
solve a linear minimisation problem of size $S$. Note that this is a
polynomial-time complexity \cite{khachian1979b}. We have that each
minimisation in Eqs.~\eqref{lms}--\eqref{ums} takes time $O(L(S))$ at most.
This holds also for the minimisation in \eqref{past} which can be converted to
a linear minimisation problem by a result from Charnes and Cooper
\cite{charnes1962}. Note that each of the mentioned minimisations must be
repeated for all the joint states of the variables in $\Pi_{A_i}^+\cap R$,
whose number is upper bounded by the states of those in $\Pi_{A_i}$. Denoting
by $H$ the worst-case number of states of the variables in $\Pi_{A_i}$
obtained by letting $i$ vary from $0$ to $m$, we have that the overall
computational complexity for Problems~\eqref{lms}--\eqref{ums} is $O(H\cdot L(S))$ at most. We can regard this part as a pre-processing step of the test of credal dominance. Once the pre-processing is over, the set of minimisations in Eqs.~\eqref{eq:ccd3.3}--\eqref{eq:ccd3.2} takes linear time in the size of $B^+$ as in the case of Bayesian networks.
\par
Case~\ref{masses} presents a lower \emph{overall} complexity for testing credal dominance. In
fact, the minimisations in Eqs.~\eqref{past}--\eqref{ums} can be solved simply
by enumerating the mass functions that make up each credal set. These mass
functions are specified directly by the modeller, i.e., they are an input of
the problem. For this reason the overall complexity of testing credal
dominance is linear in the size of $B^+$.
\par
The final case of probability intervals is also easily solved. With respect to
Eqs.~\eqref{lms}--\eqref{ums}, $\lms_0(a_i\vert \pi_{A_i}')$ and
$\ums_0(a_i\vert \pi_{A_i}'')$ are just the left and the right extreme of the
probability intervals for $(a_i\vert \pi_{A_i}')$ and $(a_i\vert
\pi_{A_i}'')$, respectively, so no computation is needed for them. As far as
Eq.~\eqref{past} is concerned, we have that the minimum of
$p_0(c'\vert\pi_C)/p_0(c''\vert\pi_C)$ taken with respect to the mass
functions in $\solp(\lpr_0^{C\vert\pi_{C}})$ is equal to
$\lms_0(c'\vert\pi_C)/\ums_0(c''\vert\pi_C)$ by the property mentioned at the
end of Case~\ref{intervals}. Again, $\lms_0(c'\vert\pi_C)$ and
$\ums_0(c''\vert\pi_C)$ are readily available as an input of the problem.
Overall, the complexity of testing credal dominance is linear in the size of $B^+$ in this case
as well.
\par
So far we have treated the case when $B^+$ is singly connected. The
extension to the general case is completely analogous to that already developed
for Bayesian networks, basically because the arcs leaving evidence nodes can be
dropped in credal networks, too. The reason is that a credal net can be
regarded as a set of Bayesian nets, and the mentioned property applies to all
the Bayesian nets in the set. More precisely, assume, as in the description at
the end of Section~\ref{sec:algorithm}, that a loop cutset is given that
together with $E$ can open all the loops in $B^+$. Call $R_1$ the loop
cutset, and let $R_2$ be the set of nodes such that $R=R_1\cup R_2$. Re-write
the test of credal dominance for credal networks as
\begin{align}
&\min_{p_0\in\solp(\lpr_0)}\min_{r\in\mathcal{R}}
\frac{p_0(c'\vert e,r)}{p_0(c''\vert e,r)}\notag\\
&=\min_{p_0\in\solp(\lpr_0)}
\left\{\min_{r_1\in\mathcal{R}_1}\left[\min_{r_2\in\mathcal{R}_2}
\frac{p_0(c'\vert e,r_1,r_2)}{p_0(c''\vert e,r_1,r_2)}\right]\right\}
\label{eq:ccd6}\\
&=\min_{r_1\in\mathcal{R}_1}
\left\{\min_{p_0\in\solp(\lpr_0)}
\left[\min_{r_2\in\mathcal{R}_2}
\frac{p_0(c'\vert e,r_1,r_2)}{p_0(c''\vert e,r_1,r_2)}
\right]\right\}.\label{eq:ccd7}
\end{align}
Eq.~\eqref{eq:ccd6} makes it clear that for each selected mass function
$p_0\in\solp(\lpr_0)$, the minimum in square brackets can be obtained on the
graph $B^+$ that is made singly connected by dropping the arcs leaving $E\cup
R_1$.  Of course this property continues to hold in the next expression. When
we consider the part in braces in \eqref{eq:ccd7}, that is, also the
variations of $p_0$, we are focusing on the singly connected credal net, with
graph $B^+$, obtained from the multiply connected one dropping the arcs
leaving $E\cup R_1$. Hence, Expression~\eqref{eq:ccd7} shows that the inner
double minimisation can be computed by Theorem~\ref{theo:credal-algorithm}.
The outer minimisation is the usual enumeration of the states of the loop
cutset.
\par
It turns out that the complexity of testing credal dominance when $B^+$ is
multiply connected is the same both for credal and Bayesian networks.  This is
an important result, as the complexity to work with credal networks is usually
much harder than that needed with Bayesian nets.

\section{Conclusions}
\label{sec:conclusions}
It seems to us that updating probabilities with incomplete observations
presents an important problem for research in uncertain reasoning, and is a
pervasive issue in applications. It has been clearly pointed out in the
literature that the commonly used CAR assumption about the incompleteness
mechanism is often unjustified, and more generally, that it may happen in
practical applications that little or no knowledge about the incompleteness
mechanism is available. In those cases, naive updating is simply
inappropriate.
\par
This paper has addressed the problem of updating probabilities when strong
assumptions about the incompleteness mechanism cannot be justified, thus
filling an important gap in literature. It has done so by deliberately
choosing the conservative point of view of not assuming any knowledge about
the incompleteness mechanism. A new so-called conservative updating method
follows as a logical consequence, using only arguments of coherence. We used
it to derive a new coherent updating rule for probabilistic expert systems. By
focusing on expert systems based on Bayesian nets, we have shown that this
conservative updating leads to efficient classification of new evidence for a
wide class of networks, so the new developments can be exploited immediately
in real environments. Furthermore, the related algorithm can be implemented
easily and does not require changes in pre-existing knowledge bases, so that
existing expert systems can be upgraded to make our robust, conservative,
inferences with minimal changes.
\par
We want to stress here that the proposed conservative updating strategy is
different in one important respect from the more traditional ones: it
generally leads only to partially determined inferences and decisions, and
ultimately to systems that can recognise the limits of their knowledge, and
suspend judgement when these limits are reached.  As necessary consequences of
our refusal to make unwarranted assumptions, we believe that these limitations
are important characteristics of the way systems ought to operate in the real
world. A system that, in a certain state, cannot support any decision on the
basis of its knowledge base, will induce a user to look for further sources of
information external to the system. In contrast, systems that may make
arbitrary choices without making that evident, will wrongly lead a user to
think that also these choices are well motivated.
\par
We also believe it is important to stress here that it is difficult to avoid
partial indeterminacy in real applications. Realistic states of partial
knowledge about the incompleteness mechanism, other than the total ignorance
modelled here, should in principle also be modelled by a (non-vacuous)
coherent lower prevision, which may again lead to indeterminacy except in very
special cases, such as when enough information is available to justify
modelling the incompleteness mechanism by a precise probability model.  For
analogous reasons, domain knowledge should most likely be modelled by a
coherent lower prevision, too. In practise this can be done by moving from
Bayesian to credal networks. It appears that this step has not really been
taken so far, probably because of the computational complexity of working in
the more general framework of credal networks. This paper shows that the
classification complexity is unchanged by moving from Bayesian to credal
networks, in the realistic scenarios that involve a state of ignorance about
the incompleteness mechanism. We hope that this encouraging result may
contribute to credal networks receiving due credit also as practical modelling
tools.
\par
With respect to future research, we believe an important issue is the
development of models able to take advantage of intermediate states of
knowledge about the incompleteness mechanism, to the extent of making stronger
inferences and decisions. With regard to Bayesian and credal nets, one could
for instance think of partitioning the set of attributes in those for which
MAR holds and the rest for which the mechanism is unknown. Such hybrid
modelling seems to provide a good compromise between generality and
flexibility.

\section*{Acknowledgements}
The authors are grateful to Peter Walley for initial stimulating discussions
on the topic of the paper in August 2001. They would also like to thank two
anonymous referees for their help in making this paper more readable, and for
pointing out a mistaken claim about computational complexity which is now
corrected.
\par
This research is partially supported by research grant G.0139.01 of the
Flemish Fund for Scientific Research (FWO), and by the Swiss NSF grant
2100-067961.

\appendix
\section{Extending Walley's Marginal Extension Theorem}
\label{sec:ext-mar-ext}
This appendix is devoted to the proof of an important theorem, needed in
Section~\ref{sec:missing-data}. It is a generalisation to three random
variables of Walley's Marginal Extension Theorem, discussed in
Section~\ref{sec:marginal-extension} (see
Theorem~\ref{theo:marginal-extension}). Because the proof is rather technical,
and it uses results and notions not explained in the main text, we have
decided to discuss it separately.
\par
We consider three random variables $\rvx$, $\rvy$ and $\rvz$ taking values in
the respective non-empty and finite spaces $\xvalues$, $\yvalues$ and
$\zvalues$.

\begin{theorem}
\label{theo:ext-mar-ext}
Consider a coherent lower prevision $\lpr$ on $\gambles(\xvalues)$, a
separately coherent conditional lower prevision $\lpr(\cdot\vert\rvx)$ on
$\gambles(\xyvalues)$, and a separately coherent conditional lower prevision
$\lpr(\cdot\vert\rvx,\rvy)$ on $\gambles(\xyzvalues)$. Then the smallest
coherent lower prevision on $\gambles(\xyzvalues)$ that has marginal $\lpr$
and is jointly coherent with $\lpr(\cdot\vert\rvx)$ and
$\lpr(\cdot\vert\rvx,\rvy)$, is given by
\begin{equation}
\label{eq:ext-mar-ext}
\alpr(h)=\lpr(\lpr(\lpr(h\vert\rvx,\rvy)\vert\rvx))
\end{equation}
for all gambles $h$ on $\xyzvalues$.
\end{theorem}

\begin{proof}
  Lemma~\ref{lem:step-one} tells us that $\alpr$ is a indeed a coherent lower
  prevision that has marginal $\lpr$. To prove that $\alpr$,
  $\lpr(\cdot\vert\rvx)$ and $\lpr(\cdot\vert\rvx,\rvy)$ are jointly coherent,
  Walley's Reduction Theorem \cite[Theorem~7.1.5]{walley1991} tells us that we
  need only prove that $\alpr$, $\lpr(\cdot\vert\rvx)$ and
  $\lpr(\cdot\vert\rvx,\rvy)$ are weakly coherent, and that
  $\lpr(\cdot\vert\rvx)$ and $\lpr(\cdot\vert\rvx,\rvy)$ are jointly coherent.
  This is done in Lemmas~\ref{lem:step-two} and~\ref{lem:step-three},
  respectively.  Finally, in Lemma~\ref{lem:step-four} we prove that any other
  coherent lower prevision on $\gambles(\xyzvalues)$ that has marginal $\lpr$
  and is jointly coherent with $\lpr(\cdot\vert\rvx)$ and
  $\lpr(\cdot\vert\rvx,\rvy)$, dominates $\alpr$.
\end{proof}

\begin{lemma}
\label{lem:step-one}
The lower prevision $\alpr$ defined on $\gambles(\xyzvalues)$ by
Eq.~\eqref{eq:ext-mar-ext} is coherent and has marginal $\lpr$.
\end{lemma}

\begin{proof}
  It is easily verified that $\alpr$ satisfies the axioms $(\lpr1)$--$(\lpr3)$
  of a coherent lower prevision, because the coherent $\lpr$, and the
  separately coherent $\lpr(\cdot\vert\rvx)$ and $\lpr(\cdot\vert\rvx,\rvy)$
  do so. It remains to show that $\alpr$ has marginal $\lpr$. Consider any
  gamble $f$ on $\xvalues$. If follows from the separate coherence of
  $\lpr(\cdot\vert\rvx,\rvy)$ that $\lpr(f\vert\rvx,\rvy)=f$ and consequently,
  from the separate coherence of $\lpr(\cdot\vert\rvx)$ that
  $\lpr(\lpr(f\vert\rvx,\rvy)\vert\rvx)=\lpr(f\vert\rvx)=f$, whence indeed
  $\alpr(f)=\lpr(\lpr(\lpr(f\vert\rvx,\rvy)\vert\rvx))=\lpr(f)$.
\end{proof}

\begin{lemma}
\label{lem:step-two}
$\alpr$, $\lpr(\cdot\vert\rvx)$ and $\lpr(\cdot\vert\rvx,\rvy)$ are weakly
coherent.
\end{lemma}

\begin{proof}
  Following the discussion in \cite[Section~7.1.4]{walley1991}, we must prove
  that
\begin{enumerate}[(a)]
\item $\max[G(f)+G(g\vert\rvx)+G(h\vert\rvx,\rvy)-G(f_0)]\geq0$;
\item $\max[G(f)+G(g\vert\rvx)+G(h\vert\rvx,\rvy)-G(g_0\vert x_0)]\geq0$;
\item $\max[G(f)+G(g\vert\rvx)+G(h\vert\rvx,\rvy)-G(h_0\vert x_0,y_0)]\geq0$;
\end{enumerate}
for all $f$, $f_0$, $h$, $h_0$ in $\gambles(\xyzvalues)$, all $g$, $g_0$ in
$\gambles(\xyvalues)$, all $x_0$ in $\xvalues$ and all $y_0$ in $\yvalues$,
where we use the notations $G(f)=f-\alpr(f)$,
$G(g\vert\rvx)=g-\lpr(g\vert\rvx)$,
$G(h\vert\rvx,\rvy)=h-\lpr(h\vert\rvx,\rvy)$, $G(g_0\vert
x_0)=I_{\{x_0\}}[g-\lpr(g\vert x_0)]$ and $G(h_0\vert
x_0,y_0)=I_{\{(x_0,y_0)\}}[h_0-\lpr(h_0\vert x_0,y_0)]$.
\par
To prove that (a) holds, recall from Lemma~\ref{lem:step-one} that $\alpr$ is
a coherent lower prevision, whence (see for instance
\cite[Section~2.6.1]{walley1991} for properties of coherent lower previsions)
\begin{multline*}
  \max[G(f)+G(g\vert\rvx)+G(h\vert\rvx,\rvy)-G(f_0)]\\
\begin{aligned}
  &\geq\aupr(G(f)+G(g\vert\rvx)+G(h\vert\rvx,\rvy)-G(f_0))\\
  &\geq\alpr(G(f)+G(g\vert\rvx)+G(h\vert\rvx,\rvy))
  -\alpr(G(f_0))\\
  &\geq\alpr(G(f))+\alpr(G(g\vert\rvx))+\alpr(G(h\vert\rvx,\rvy))
  -\alpr(G(f_0))
\end{aligned}
\end{multline*}
Now, again using the coherence of $\alpr$, we find that
$\alpr(G(f))=\alpr(f-\alpr(f))=\alpr(f)-\alpr(f)=0$ and similarly
$\alpr(G(f_0))=0$. Moreover, it follows from the separate coherence of
$\lpr(\cdot\vert\rvx,\rvy)$ that for all $(x,y)$ in $\xyvalues$
\begin{multline*}
  \lpr(G(h\vert\rvx,\rvy)\vert x,y)
  =\lpr(h-\lpr(h\vert\rvx,\rvy)\vert x,y)\\
  =\lpr(h-\lpr(h\vert x,y)\vert x,y) =\lpr(h\vert x,y)-\lpr(h\vert x,y) =0,
\end{multline*}
whence $\lpr(G(h\vert\rvx,\rvy)\vert\rvx,\rvy)=0$ and consequently
$\alpr(G(h\vert\rvx,\rvy))=0$. Similarly, it follows from the separate
coherence of $\lpr(\cdot\vert\rvx,\rvy)$ that
$\lpr(G(g\vert\rvx)\vert\rvx,\rvy)=G(g\vert\rvx)$, and from the separate
coherence of $\lpr(\cdot\vert\rvx)$ that for all $x$ in $\xvalues$,
\begin{multline*}
  \lpr(\lpr(G(g\vert\rvx)\vert\rvx,\rvy)\vert x) =\lpr(G(g\vert\rvx)\vert x)
  =\lpr(g-\lpr(g\vert\rvx)\vert x)\\
  =\lpr(g-\lpr(g\vert x)\vert x) =\lpr(g\vert x)-\lpr(g\vert x)=0,
\end{multline*}
whence $\lpr(\lpr(G(g\vert\rvx)\vert\rvx,\rvy)\vert\rvx)=0$ and consequently
also $\alpr(G(g\vert\rvx))=0$. It follows that (a) is indeed verified.
\par
An argument similar to the one above tells us that (b) will hold if we can
prove that $\alpr(G(g_0\vert x_0))=0$. Now it follows from the separate
coherence of $\lpr(\cdot\vert\rvx,\rvy)$ that, since $G(g_o\vert
x_0)\in\gambles(\xyvalues)$, $\lpr(G(g_o\vert x_0)\vert\rvx,\rvy)=G(g_o\vert
x_0)$, whence, using the separate coherence of $\lpr(\cdot\vert\rvx)$,
\begin{align*}
  \lpr(\lpr(G(g_o\vert x_0)\vert\rvx,\rvy)\vert\rvx)
  &=\lpr(G(g_o\vert x_0)\vert\rvx)\\
  &=\lpr(I_{\{x_0\}}[g(x_0,\cdot)-\lpr(g_o(x_0,\cdot)\vert x_0)]\vert\rvx)\\
  &=I_{\{x_0\}}[\lpr(g(x_0,\cdot)\vert\rvx)-\lpr(g_o(x_0,\cdot)\vert x_0)] =0,
\end{align*}
whence indeed $\alpr(G(g_0\vert x_0))=0$.
\par
Similarly, (c) will be verified if we can prove that $\alpr(G(h_0\vert
x_0,y_0))=0$.  Now it follows from the separate coherence of
$\lpr(\cdot\vert\rvx,\rvy)$ that
\begin{multline*}
  \lpr(G(h_0\vert x_0,y_0)\vert\rvx,\rvy)
  =\lpr(I_{\{(x_0,y_0)\}}[h-\lpr(h\vert x_0,y_0)]\vert\rvx,\rvy)\\
  =I_{\{(x_0,y_0)\}}[\lpr(h\vert\rvx,\rvy)-\lpr(h\vert x_0,y_0)]=0,
\end{multline*}
whence indeed $\alpr(G(h_0\vert x_0,y_0))=0$.
\end{proof}

\begin{lemma}
\label{lem:step-three}
Separately coherent conditional lower previsions $\lpr(\cdot\vert\rvx)$ on
$\gambles(\xyvalues)$ and $\lpr(\cdot\vert\rvx,\rvy)$ on
$\gambles(\xyzvalues)$ are always jointly coherent.
\end{lemma}

\begin{proof}
  We use the discussion of joint coherence in
  \cite[Section~7.1.4]{walley1991}. Consider arbitrary $g$ in
  $\gambles(\xyvalues)$ and $h$ in $\gambles(\xyzvalues)$ and the
  corresponding sets
\begin{equation*}
S(g)=\set{\{x\}\times\yvalues\times\zvalues}{g(x,\cdot)\not=0}
\text{ and }
S(h)=\set{\{x\}\times\{y\}\times\zvalues}{h(x,y,\cdot)\not=0}.
\end{equation*}
First of all, consider any $x_0$ in $\xvalues$ and $g_0$ in
$\gambles(\xyvalues)$, then we must show that there is some $B$ in
\begin{equation*}
S(g)\cup S(h)\cup\left\{\{x_0\}\times\yvalues\times\zvalues\right\}
\end{equation*}
such that (if we also take into account the separate coherence of
$\lpr(\cdot\vert\rvx,\rvy)$ and $\lpr(\cdot\vert\rvx)$)
\begin{multline*}
  \max_{(x,y,z)\in B} \big[ g(x,y)-\lpr(g(x,\cdot)\vert x)
  +h(x,y,z)-\lpr(h(x,y,\cdot)\vert x,y)\\
  -I_{\{x_0\}}(x)(g_0(x,y)-\lpr(g_0(x,\cdot)\vert x)) \big] \geq0
\end{multline*}
We choose $B=\{x_0\}\times\yvalues\times\zvalues$, and prove that the
corresponding supremum
\begin{multline*}
  S=\max_{y\in\yvalues}\max_{z\in\zvalues} \big[
  g(x_0,y)-\lpr(g(x_0,\cdot)\vert x_0)
  +h(x_0,y,z)-\lpr(h(x_0,y,\cdot)\vert x_0,y)\\
  -(g_0(x_0,y)-\lpr(g_0(x_0,\cdot)\vert x_0)) \big]\geq0.
\end{multline*}
Now, since it follows from the coherence of the lower prevision
$\lpr(\cdot\vert x_0,y)$ that
\begin{equation*}
\max_{z\in\zvalues}
\left[
h(x_0,y,z)-\lpr(h(x_0,y,\cdot)\vert x_0,y)
\right]
\geq0
\end{equation*}
for all $y\in\yvalues$, we see that indeed
\begin{align*}
  S\geq\max_{y\in\yvalues} \left[g(x_0,y)-\lpr(g(x_0,\cdot)\vert x_0)
    -(g_0(x_0,y)-\lpr(g_0(x_0,\cdot)\vert x_0)) \right]\geq0,
\end{align*}
where the last inequality follows from the coherence of the lower prevision
$\lpr(\cdot\vert x_0)$.
\par
As a second step, consider any $(x_0,y_0)$ in $\xyvalues$ and $h_0$ in
$\gambles(\xyzvalues)$, then we must show that there is some $B$ in
\begin{equation*}
S(g)\cup S(h)\cup\left\{\{x_0\}\times\{y_0\}\times\zvalues\right\}
\end{equation*}
such that (if we also take into account the separate coherence of
$\lpr(\cdot\vert\rvx,\rvy)$ and $\lpr(\cdot\vert\rvx)$)
\begin{multline*}
  \max_{(x,y,z)\in B} \big[ g(x,y)-\lpr(g(x,\cdot)\vert x)
  +h(x,y,z)-\lpr(h(x,y,\cdot)\vert x,y)\\
  -I_{\{(x_0,y_0)\}}(x,y)(h_0(x,y,z)-\lpr(h_0(x,y,\cdot)\vert x,y)) \big]
  \geq0
\end{multline*}
If $g(x_1,\cdot)\not=0$ for some $x_1\not=x_0$, then we choose
$B=\{x_1\}\times\yvalues\times\zvalues$, and similar arguments as in the first
step of the proof lead us to conclude that the corresponding supremum
\begin{equation*}
\max_{y\in\yvalues}\max_{z\in\zvalues}
\big[
g(x_1,y)-\lpr(g(x_1,\cdot)\vert x_1)
+h(x_1,y,z)-\lpr(h(x_1,y,\cdot)\vert x_1,y)
\big]
\end{equation*}
is indeed non-negative. Assume therefore that $g(x,\cdot)=0$ for all
$x\not=x_0$. Then there are two possibilities left. Either $g(x_0,\cdot)=0$,
whence $g=0$. Then we choose $B=\{x_0\}\times\{y_0\}\times\zvalues$, and it
follows from the coherence of the lower prevision $\lpr(\cdot\vert x_0,y_0)$
that for the corresponding supremum;
\begin{equation*}
\max_{z\in\zvalues}
\big[
h(x_0,y_0,z)-\lpr(h(x_0,y_0,\cdot)\vert x_0,y_0)
-(h_0(x_0,y_0,z)-\lpr(h_0(x_0,y_0,\cdot)\vert x_0,y_0))
\big]
\geq0.
\end{equation*}
Or $g(x_0)\not=0$ and then we choose $B=\{x_0\}\times\yvalues\times\zvalues$,
and it follows, in a similar way as in the first step of the proof, that the
corresponding supremum
\begin{multline*}
  \max_{y\in\yvalues,z\in\zvalues} \big[ g(x_0,y)-\lpr(g(x_0,\cdot)\vert x_0)
  +h(x_0,y,z)-\lpr(h(x_0,y,\cdot)\vert x_0,y)\\
  -I_{\{y_0\}}(y)(h_0(x_0,y,z)-\lpr(h_0(x_0,y,\cdot)\vert x_0,y)) \big]
\end{multline*}
is again non-negative.
\end{proof}

\begin{lemma}
\label{lem:step-four}
Any coherent lower prevision $\alpr'$ on $\gambles(\xyzvalues)$ that has
marginal $\lpr$ and is jointly coherent with $\lpr(\cdot\vert\rvx)$ and
$\lpr(\cdot\vert\rvx,\rvy)$, dominates $\alpr$.
\end{lemma}

\begin{proof}
  Consider any $h$ in $\gambles(\xyzvalues)$, then we have to prove that
  $\alpr'(h)\geq\alpr(h)$.  Since $\alpr'$ jointly coherent with
  $\lpr(\cdot\vert\rvx)$ and $\lpr(\cdot\vert\rvx,\rvy)$, it follows that
  $\alpr'$, $\lpr(\cdot\vert\rvx)$ and $\lpr(\cdot\vert\rvx,\rvy)$ are weakly
  coherent (see \cite[Section~7.1.4]{walley1991}), and consequently we have
  for any $h_0$, $h_1$ and $g$ in $\gambles(\xyzvalues)$, and any $f$ in
  $\gambles(\xyvalues)$ that
\begin{equation*}
\max\left[
h_1-\alpr'(h_1)+f-\lpr(f\vert\rvx)+g-\lpr(g\vert\rvx,\rvy)
-(h_0-\alpr'(h_0))
\right]\geq0.
\end{equation*}
If we choose $h_0=g=h$, $f=\lpr(h\vert\rvx,\rvy)$ and
$h_1=\lpr(\lpr(h\vert\rvx,\rvy)\vert\rvx)$, this reduces to
\begin{equation*}
\alpr'(h)\geq\alpr'(\lpr(\lpr(h\vert\rvx,\rvy)\vert\rvx))
\end{equation*}
and since $\lpr(\lpr(h\vert\rvx,\rvy)\vert\rvx)$ is a gamble on $\xvalues$,
and $\alpr'$ has marginal $\lpr$, we find that
\begin{equation*}
\alpr'(\lpr(\lpr(h\vert\rvx,\rvy)\vert\rvx))
=\lpr(\lpr(\lpr(h\vert\rvx,\rvy)\vert\rvx))
=\alpr(h),
\end{equation*}
whence indeed $\alpr'(h)\geq\alpr(h)$.
\end{proof}

\section{Additional discussion of the irrelevance
  condition~\eqref{eq:md-irrelevance}}
\label{sec:extra-independence}
This appendix provides additional discussion of the irrelevance
assumption~\eqref{eq:md-irrelevance} in Section~\ref{sec:missing-data}. We use
the notations established there. We shall restrict ourselves to the case that
the lower prevision $\lpr_0$ and the conditional lower prevision
$\lpr_0(\cdot\vert\srv)$ are precise.
\par
It turns out that if we make Assumption~\eqref{eq:md-irrelevance}, coherence
guarantees that another type of irrelevance is satisfied, as the following
theorem makes clear.

\begin{theorem}
\label{theo:md-extra-independence}
Assume we have a linear prevision $\pr_0$ on $\gambles(\states)$, and a linear
conditional prevision $\pr_0(\cdot\vert\srv)$ on
$\gambles(\classes\times\states)$. Also assume that the irrelevance
condition~\eqref{eq:md-irrelevance} holds.  Then for all $x$ in $\states$ and
$c$ in $\classes$ such that $p_0(c,x)=p_0(x)p_0(c\vert x)>0$, and for all
gambles $f$ on $\observs$, the conditional lower prevision $\lpr(f\vert c,x)$
is uniquely determined by coherence, and given by
\begin{equation*}
\lpr(f\vert c,x)=\lpr(f\vert x)=\min_{o\in\mvm(x)}f(o).
\end{equation*}
\end{theorem}

\begin{proof}
  Let us first consider $\lpr(\{c\}\times\{x\}\times\observs)$. For any
  $y\in\states$ and $p\in\observs$, we have, by separate coherence, that
\begin{equation*}
\lpr(\{c\}\times\{x\}\times\observs\vert y,p)
=\lpr(I_{\{c\}\times\{x\}\times\observs}(\cdot,y,p)\vert y,p)
=I_{\{x\}}(y)\lpr(\{c\}\vert y,p),
\end{equation*}
whence $\lpr(\{c\}\times\{x\}\times\observs\vert\srv,\orv)
=I_{\{x\}}\lpr(\{c\}\vert\srv,\orv)$. Consequently, for all $y\in\states$,
using separate coherence, Eq.~\eqref{eq:md-missingness} and the irrelevance
condition~\eqref{eq:md-irrelevance},
\begin{multline*}
  \lpr(\lpr(\{c\}\times\{x\}\times\observs\vert X,O)\vert y)
  =\lpr(I_{\{x\}}(y)\lpr(\{c\}\vert y,\orv)\vert y)\\
  =I_{\{x\}}(y)\min_{p\in\mvm(x)}\lpr(\{c\}\vert x,p)
  =I_{\{x\}}(y)\min_{p\in\mvm(x)}\pr_0(\{c\}\vert x) =I_{\{x\}}(y)p_0(c\vert
  x),
\end{multline*}
whence $\lpr(\lpr(\{c\}\times\{x\}\times\observs\vert X,O)\vert X)
=I_{\{x\}}p_0(c\vert x)$, and therefore,
\begin{multline*}
  \lpr(\{c\}\times\{x\}\times\observs)
  =\pr_0(\lpr(\lpr(\{c\}\times\{x\}\times\observs\vert X,O)\vert X))\\
  =\pr_0(I_{\{x\}}p_0(c\vert x)) =p_0(x)p_0(c\vert x)) =p_0(c,x).
\end{multline*}
The material in Section~\ref{sec:joint-coherence} then tells us that whenever
$\lpr(\{c\}\times\{x\}\times\observs)=p_0(c,x)>0$, $\lpr(f\vert c,x)$ is
uniquely determined by coherence as the unique solution of the following
equation in $\mu$:
\begin{equation}
\label{eq:md-extra-gbr}
\lpr(I_{\{c\}\times\{x\}\times\observs}[f-\mu])=0.
\end{equation}
Now, for any $y\in\states$ and $p\in\observs$, we have, by separate coherence,
that
\begin{multline*}
  \lpr(I_{\{c\}\times\{x\}\times\observs}[f-\mu]\vert y,p)
  =\lpr(I_{\{c\}\times\{x\}\times\observs}(\cdot,y,p)[f-\mu]\vert y,p)\\
  =I_{\{x\}}(y)\lpr(I_{\{c\}}[f-\mu]\vert y,p),
\end{multline*}
whence $\lpr(I_{\{c\}\times\{x\}\times\observs}[f-\mu]\vert\srv,\orv)
=I_{\{x\}}\lpr(I_{\{c\}}[f-\mu]\vert\srv,\orv)$. Consequently, for all
$y\in\states$, using separate coherence, Eq.~\eqref{eq:md-missingness} and the
irrelevance assumption~\eqref{eq:md-irrelevance},
\begin{multline*}
  \lpr(\lpr(I_{\{c\}\times\{x\}\times\observs}
  [f-\mu]\vert\srv,\orv)\vert y)\\
\begin{aligned}
  &=\lpr(I_{\{x\}}(y)\lpr(I_{\{c\}}[f-\mu]\vert y,\orv)\vert y)
  =I_{\{x\}}(y)\min_{o\in\mvm(x)}\lpr(I_{\{c\}}[f(o)-\mu]\vert x,o)\\
  &=I_{\{x\}}(y)\min_{o\in\mvm(x)}\pr_0(I_{\{c\}}[f(o)-\mu]\vert x)
  =I_{\{x\}}(y)\min_{o\in\mvm(x)}[f(o)-\mu]p_0(c\vert x)\\
  &=I_{\{x\}}(y)p_0(c\vert x) \left[\min_{o\in\mvm(x)}f(o)-\mu\right]
  =I_{\{x\}}(y)p_0(c\vert x) \left[\lpr(f\vert x)-\mu\right],
\end{aligned}
\end{multline*}
whence $\lpr(\lpr(I_{\{c\}\times\{x\}\times\observs}\vert X,O)\vert X)
=I_{\{x\}}p_0(c\vert x)[\lpr(f\vert x)-\mu]$, and therefore,
\begin{multline*}
  \lpr(I{\{c\}\times\{x\}\times\observs}[f-\mu])\\
\begin{aligned}
  &=\pr_0(\lpr(\lpr(I_{\{c\}\times\{x\}\times\observs}
  [f-\mu]\vert X,O)\vert X))\\
  &=\pr_0(I_{\{x\}}p_0(c\vert x)[\lpr(f\vert x)-\mu])
  =p_0(x)p_0(c\vert x)[\lpr(f\vert x)-\mu]\\
  &=p_0(c,x)[\lpr(f\vert x)-\mu].
\end{aligned}
\end{multline*}
If $p_0(c,x)>0$, it follows that the unique solution of
Eq.~\eqref{eq:md-extra-gbr} is indeed given by $\mu=\lpr(f\vert x)$.
\end{proof}

This theorem tells us that for a linear prior $\pr_0$, the irrelevance
assumption~\eqref{eq:md-irrelevance} implies, through arguments of coherence,
that conditional on the attributes $\srv$, the class $\crv$ is irrelevant to
the observations $\orv$, i.e., if we know that $\srv=x$, then the additional
knowledge that $\crv=c$ does not change our beliefs about the value of $\orv$.
\par
We now intend to show that the above statement does not imply \eqref{eq:md-irrelevance}.  Let us, to this effect,
start with a linear prevision $\pr_0$ on $\gambles(\classes\times\states)$,
and assume that for all gambles $f$ on $\observs$, and all $(c,x)$ in
$\classes\times\states$ such that $p_0(c,x)=p_0(x)p_0(c\vert x)>0$:
\begin{equation}
\label{eq:md-alternative-irrelevance}
\lpr(f\vert c,x)=\lpr(f\vert x)=\min_{o\in\mvm(x)}f(o).
\tag{I'}
\end{equation}
We can now use Walley's marginal extension theorem (see
Theorem~\ref{theo:marginal-extension} in Section~\ref{sec:marginal-extension})
to combine the marginal linear prevision $\pr_0$ on
$\gambles(\classes\times\states)$ and the conditional lower prevision
$\lpr(\cdot\vert\crv,\srv)$ on $\gambles(\observs)$---or, through separate
coherence, on $\gambles(\classes\times\states\times\observs)$---into a joint
lower prevision $\alpr$ on $\gambles(\classes\times\states\times\observs)$
defined by
\begin{equation*}
\alpr(h)=\pr_0(\lpr(h\vert\crv,\srv))
\end{equation*}
for all gambles $h$ on $\classes\times\states\times\observs$. The following
theorem tells us that Assumption~\eqref{eq:md-irrelevance} is effectively
stronger than Assumption~\eqref{eq:md-alternative-irrelevance}.

\begin{theorem}
\label{theo:md-alternative-irrelevance}
Assume that ~\eqref{eq:md-alternative-irrelevance} holds.  Consider a
separately coherent conditional lower prevision $\lpr(\cdot\vert\srv,\orv)$ on
$\gambles(\classes\times\states\times\observs)$. If this conditional lower
prevision satisfies~\eqref{eq:md-irrelevance}, i.e.,
\begin{equation*}
\lpr(f\vert x,o)=\pr_0(f\vert x)
\end{equation*}
for all $f\in\gambles(\classes)$ , for all $x\in\states$ such that $p_0(x)>0$,
and for all $o\in\mvm(x)$, then it cannot be jointly coherent with the joint
lower prevision $\alpr$ on $\gambles(\classes\times\states\times\observs)$.
\end{theorem}

\begin{proof}
  Let $x\in\states$ such that $p_0(x)>0$ and let $o\in\mvm(x)$.  Consider an
  arbitrary gamble $f$ on $\classes$ that is not almost everywhere constant on
  $\classes$ with respect to the linear prevision $\pr_0(\cdot\vert x)$ (which
  is uniquely determined from $\pr_0$ through coherence). The theorem is
  proved if we can show that
\begin{equation*}
\alpr([f-\lpr(f\vert x,o)]
I_{\classes\times\{x\}\times\{o\}})<0.
\end{equation*}
By separate coherence and Assumption~\eqref{eq:md-alternative-irrelevance}, we
find for any $c\in\classes$ and $y\in\states$ that
\begin{multline*}
  \lpr([f-\lpr(f\vert x,o)]
  I_{\classes\times\{x\}\times\{o\}})\vert c,y)\\
\begin{aligned}
  &=\lpr([f(c)-\lpr(f\vert x,o)]
  I_{\classes\times\{x\}\times\{o\}}(c,y,\cdot))\vert c,y)\\
  &=I_{\{x\}}(y)\min_{p\in\mvm(x)}[f(c)-\lpr(f\vert x,o)]I_{\{o\}}(p)\\
  &=I_{\{x\}}(y)I_{\{o\}^*}(x)\min\{f(c)-\lpr(f\vert x,o),0\}\\
  &=I_{\{x\}}(y)I_{\{o\}^*}(x)\min\{f(c)-\pr_0(f\vert x),0\}
\end{aligned}
\end{multline*}
where the last equality follows from the assumptions of the theorem.
Consequently,
\begin{equation*}
\lpr([f-\lpr(f\vert x,o)]
I_{\classes\times\{x\}\times\{o\}})\vert\crv,\srv)
=I_{\{x\}}I_{\{o\}^*}(x)\min\{f-\pr_0(f\vert x),0\}
\end{equation*}
and we find that
\begin{multline*}
  \alpr([f-\lpr(f\vert x,o)]
  I_{\classes\times\{x\}\times\{o\}})\\
\begin{aligned}
  &=\pr_0(\lpr([f-\lpr(f\vert x,o)]
  I_{\classes\times\{x\}\times\{o\}})\vert\crv,\srv))\\
  &=\pr_0(I_{\{x\}}I_{\{o\}^*}(x)\min\{f-\pr_0(f\vert x),0\})\\
  &=I_{\{o\}^*}(x)\pr_0(I_{\{x\}}\min\{f-\pr_0(f\vert x),0\})\\
  &=I_{\{o\}^*}(x)p_0(x)\pr_0(\min\{f-\pr_0(f\vert x),0\}\vert x)<0,
\end{aligned}
\end{multline*}
where the inequality follows from $x\in\{o\}^*$, $p_0(x)>0$, and
Lemma~\ref{lem:auxiliary}.
\end{proof}

\begin{lemma}
\label{lem:auxiliary}
Let $\pr$ be a linear prevision on $\gambles(\classes)$. Then for all gambles
$f$ on $\classes$ that are not almost everywhere constant (with respect to
$\pr$), and for all real $\mu$, we have that
\begin{equation*}
\pr(\min\{f-\mu,0\})\geq0\Rightarrow\mu<\pr(f).
\end{equation*}
\end{lemma}

\begin{proof}
  Let $f$ be a gamble that is not constant almost everywhere, i.e., $f$ is not
  constant on the set $D_p=\set{c\in\classes}{p(c)>0}$, where we denote by $p$
  the mass function of $\pr$. It clearly suffices to show that
  $\pr(\min\{f-\pr(f),0\})<0$.  Assume, \emph{ex absurdo}, that
  $\pr(\min\{f-\pr(f),0\})\geq0$.  Since the gamble $\min\{f-\pr(f),0\}$ on
  $\classes$ is non-positive, this implies that $\pr(\min\{f-\pr(f),0\})=0$,
  and this can only happen if $p(c)=\pr(\{c\})=0$ for all $c\in\classes$ such
  that $f(c)<\pr(f)$.  Consequently, $\pr(f)\leq f(c)$ for all $c\in D_p$,
  whence $\pr(f)\leq\min_{c\in D_p}f(c)$. But since $\pr(f)$ is a non-trivial
  convex mixture of the $f(c)$ for all $c\in D_p$, and since $f$ is not
  constant on $D_p$, we also know that $\pr(f)>\min_{c\in D_p}f(c)$, a
  contradiction.
\end{proof}

\bibliographystyle{plain} 
\bibliography{general}

\end{document}